\documentclass[10pt,twocolumn,letterpaper]{article}

\usepackage{iccv}
\usepackage{times}
\usepackage{epsfig}
\usepackage{graphicx}
\usepackage{amsmath}
\usepackage{color}
\usepackage{amssymb}
\usepackage{lipsum}

\usepackage[numbers,sort]{natbib}
\usepackage{xcolor, colortbl}
\usepackage{mathtools}
\usepackage{tabularx}
\usepackage{multirow}
\usepackage{enumitem}
\usepackage{bbm}
\usepackage{wrapfig}
\usepackage{xspace}
\usepackage{array}

\def\ie{\emph{i.e.}}
\def\eg{\emph{e.g.}}
\def\etal{\emph{et al.}}

\definecolor{grn}{rgb}{0.1, 0.6, 0.1}
\definecolor{mgt}{rgb}{0.6, 0.1, 0.6}
\definecolor{red}{rgb}{0.7 0.1 0.1}
\definecolor{blue}{rgb}{0.0 0.0 1.0}
\definecolor{amethyst}{rgb}{0.6, 0.4, 0.8}

\newcommand{\sftype}[1]{{\textsf{\small #1}}}
\newcommand*{\mcomb}[2]{{}^{#1}C_{#2}}
\newcommand{\expnum}[2]{{#1}\mathrm{e}{-#2}}

\usepackage[pagebackref=true,breaklinks=true,letterpaper=true,colorlinks,bookmarks=false]{hyperref}

\iccvfinalcopy 

\def\iccvPaperID{3640} 

\ificcvfinal\fi

\begin{document}
\title{ASMR: Learning Attribute-Based Person Search with\\Adaptive Semantic Margin Regularizer}

\author{
Boseung Jeong$^1$\textsuperscript{*}\qquad \qquad \qquad
Jicheol Park$^2$\textsuperscript{*} \qquad \qquad \qquad
Suha Kwak$^{1,2}$ \\
Dept. of CSE, POSTECH$^1$, \qquad
Graduate School of AI, POSTECH$^2$\\
{\tt\small \url{http://cvlab.postech.ac.kr/research/ASMR/}}
}

\maketitle
\begingroup\renewcommand\thefootnote{*}
\footnotetext{Equal contribution}
\renewcommand*{\thefootnote}{\arabic{footnote}}

\begin{abstract}
Attribute-based person search is the task of finding person images that are best matched with a set of text attributes given as query.
The main challenge of this task is the large modality gap between attributes and images. 
To reduce the gap, we present a new loss for learning cross-modal embeddings in the context of attribute-based person search. 
We regard a set of attributes as a category of people sharing the same traits.
In a joint embedding space of the two modalities, our loss pulls images close to their person categories for modality alignment.
More importantly, it pushes apart a pair of person categories by a margin determined adaptively by their semantic distance, where the distance metric is learned end-to-end so that the loss considers importance of each attribute when relating person categories.
Our loss guided by the adaptive semantic margin leads to more discriminative and semantically well-arranged distributions of person images.
As a consequence, it enables a simple embedding model to achieve state-of-the-art records on public benchmarks without bells and whistles.

\end{abstract}


\section{Introduction}

Person search is the task of finding people from a large set of images given a query describing their appearances. 
It plays critical roles in applications for public safety such as searching for criminals in videos and tracking people using multiple surveillance cameras with non-overlapping fields of view.
Person search has been formulated as a fine-grained image retrieval problem focusing only on person images, where a solution should discriminate subtle appearance variations of different people 
and at the same time generalize well to people unseen during training.

Most of existing person search techniques need an image that exemplifies target person as query~
\cite{Li_2014_CVPR, Sun_2018_ECCV, Kalayeh_2018_CVPR, Zhong_2018_CVPR, Ali_2018_ECCV, He_2018_CVPR, Chen_2019_ICCV, Xiao2017CVPR, liu2017neural, chang2018rcaa, lan2018person, chen2018person, yan2019learning, park2020learning}. 
However, image query is not always accessible in real world scenarios, \eg, where eyewitness memory is the only evidence for finding criminals.
A solution to this issue is to utilize a verbal description as query for person search~\cite{li2017identity, li2017person}, but it suffers from the inherent ambiguity of natural language and requires complicated processes to understand the query.

To address the above issue, we study in this paper person search using text attributes as query. 
Specifically, a query is given as a set of predefined attributes indicating traits of target person, \eg, gender, age, clothing, and accessory; we consider such a set as a \emph{person category}, and multiple people sharing the same traits belong to the same person category.
This approach is suitable for person search in the wild since attributes are cheap to collect while being less ambiguous and more tractable than natural language descriptions.
The use of attributes as query, however, introduces additional challenges due to the limited descriptive capability of attributes, which leads to a large modality gap between images and person categories.

Previous work on attribute-based person search attempts to reduce the modality gap 
by aligning each person category and corresponding images in a joint embedding space through modality-adversarial training~\cite{yin2017adversarial,cao2020symbiotic} or 
by enhancing the expressive power of embedding vectors of person categories and images in a hierarchical manner~\cite{dong2019person}.
Although these pioneer studies shed light on the important yet less explored approach to person search, there is still large room for further improvement.
First, they are unstable and computationally heavy in training due to their adversarial learning strategies~\cite{yin2017adversarial,cao2020symbiotic}, or expensive in inference due to the large dimensional embedding vectors demanding an extra network to be matched~\cite{dong2019person}.
More importantly, these methods treat person categories as independent class labels of person images and ignore their relations, \eg, how many attributes are different between them,
although such relations can provide a rich supervisory signal for learning better representations of person categories and images.

We develop a new attribute-based person search method that overcomes these limitations. 
Our method learns a joint embedding space of the two different modalities through a pair of simple encoder networks, one for images and the other for person categories;
a person category is represented as a binary vector, each of whose dimensions indicates the presence of the corresponding attribute. 
When conducting person search, a person category is given as query in the form of binary vector and projected onto the joint embedding space by the person category encoder, then images whose embedding vectors are closest to that of the query in the space are retrieved.

The main contribution of this work is a new loss function, which enables our model to achieve outstanding performance with the simple architecture and retrieval pipeline.
In the joint embedding space, the loss regards each person category as a semantic prototype of associated images, and encourages the images to be close to their prototype so that the two modalities are aligned.
The key feature of the loss is that it determines the margin between person categories in the embedding space adaptively by their distance in the binary attribute space.
Moreover, the distance is measured by weighted Hamming metric, 
in which weights multiplied to individual bits (\ie, attributes) are optimized together with parameters of the embedding networks so that the loss focuses on more important attributes when relating person categories. 
This idea is implemented by Adaptive Semantic Margin Regularizer (ASMR) as a part of our loss.

The proposed loss function with ASMR allows the distributions of person images to be more discriminative and semantically well-arranged in the learned embedding space.
Consequently, our method achieves the state of the art on three public benchmark datasets~\cite{PETA,Market1501_attribute,PA100K} without bells and whistles.
Also, compared to the previous work~\cite{yin2017adversarial,cao2020symbiotic,dong2019person}, it is efficient since it works on an embedding space of a small dimension with no extra network, and converges very quickly in training since it does not require adversarial training.
The main contribution of our work is three-fold:
\vspace{-1mm}
\begin{itemize}[leftmargin=5mm] 
\itemsep=-0.5mm
   \item We propose a novel cross-modal embedding loss, considering semantic relations between person categories so that the embedding space becomes more discriminative and better generalizes to unseen categories. 
   \item The straightforward architecture and retrieval pipeline of the proposed framework enable fast convergence in training and efficient person search in testing.
   \item Our method achieves the state of the art on three public benchmarks without bells and whistles. 
\end{itemize}


\begin{figure*}[t!]
    \centering
    \includegraphics[width=\textwidth]{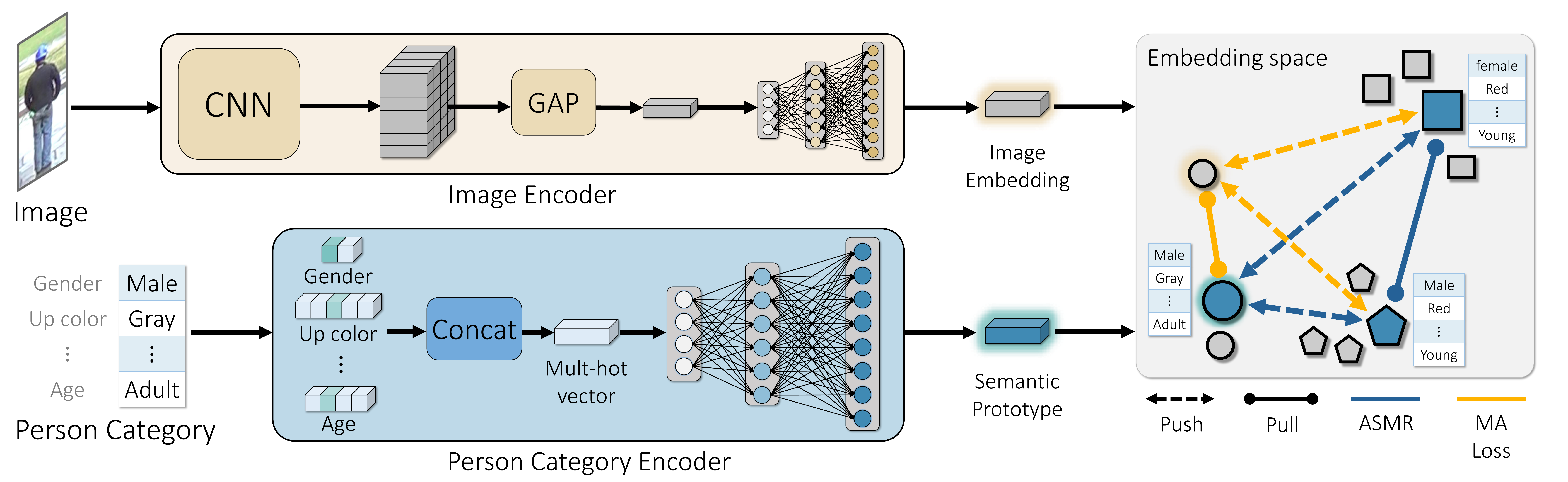}
    \caption{
    Overall pipeline of our method. Image is embedded by a conventional CNN followed by a MLP while the query set of attributes, called \emph{person category}, is converted to a binary vector and encoded through a separate embedding network. In their joint embedding space, a positive pair of image embedding and semantic prototype are pulled together while a negative pair is pushed apart for cross-modal alignment. Also, a pair of semantic prototypes pushes or pulls each other by a margin determined adaptively by their semantic affinity.
    }
    \label{fig:architecture}
    \vspace{-2mm}
\end{figure*}


\section{Related Work}

\subsection{Attribute-Based Person Search}

A na\"ive approach to attribute-based person search is recognizing attributes of person images and finding images whose predicted attributes are the same with the person category given as query~\cite{vaquero2009attribute, scheirer2012multi, li2015multi}.
However, this approach is unreliable due to imperfection of attribute recognition. 
Note that attribute recognition itself is challenging since the appearance of an attribute could vary significantly and person images captured by surveillance cameras are often limited in terms of resolution and quality. 

Recent methods instead learn and utilize a cross-modal embedding space where person categories and associated images are close to each other.
The main issue in this direction is the large gap between the two modalities.
Dong~\etal~\cite{dong2019person} 
tackle the problem by capturing rich information of the two modalities through hierarchical embeddings.
However, their model is computationally heavy since it computes high dimensional embeddings and deploys an extra network for matching them.
Yin~\etal~\cite{yin2017adversarial} and Cao~\etal~\cite{cao2020symbiotic} learn a joint embedding space where person categories and images are matched directly. 
To bridge the modality gap, their embedding spaces are trained in modality-adversarial manners, which however often result in unstable and tardy convergence due to the nature of the minimax optimization.
Moreover, these methods share a limitation that person categories are considered as individual class labels and their nontrivial relations are ignored.

Our method also learns a cross-modal embedding space, but unlike the previous arts, it is efficient in both training and testing, and lets the learned embedding space reflect semantic relations between person categories.

\subsection{Deep Metric Learning}

The goal of deep metric learning is to learn an embedding space where data of the same class are grouped together and those of different classes are pushed away. 
Loss functions for metric learning are roughly categorized into two classes, pair-based and proxy-based losses.

Pair-based losses basically pull a pair of embedding vectors close to each other if they are of the same class and push them apart otherwise. 
An early example following this principle is contrastive loss~\cite{Hadsell2006,Bromley1994,Chopra2005}, which is extended to consider higher order relations of embedding vectors by associating multiple pairs~\cite{Schroff2015,Wang2014,Sohn_nips2016,songCVPR16,wang2019multi}.
On the other hand, proxy-based losses relate embedding vectors with prototypes, each of which is a virtual embedding vector typifying each class of training data and learned as a part of embedding network.
Then the losses pull together or push apart a pair of embedding vector and prototype according to their class equivalence~\cite{movshovitz2017no, Kim_2020_CVPR, deng2019arcface}.

Unfortunately, these losses are not proper to be applied directly to attribute-based person search for the following reasons.
First, most of them are developed for uni-modal retrieval, except few examples~\cite{faghri2017vse++,liong2016deep}.
Second, they cannot take semantic relations between person categories into account since they regard the categories as independent labels whose relations are binary (\ie, the same or not).

Unlike the existing losses for metric learning, our loss can handle the nontrivial inter-category relations as well as those between categories and images thanks to ASMR.
We believe that our loss can be applied to other tasks where inter-label relations are beyond the binary.

\subsection{Cross-Modal Retrieval}
Attribute-based person search is a particular example of cross-modal retrieval, which has been studied mainly for image-text or image-sound retrieval~\cite{wang2015deep, wang2016learning, wang2017adversarial, eisenschtat2017linking, nagrani2018learnable, wang2019learning}.
Most of existing methods for cross-modal retrieval aim to learn a joint embedding space of different modalities so that a simple nearest neighbor search can find samples of the same content in the space regardless of their modalities.
This idea has been implemented in general by Canonical Correlation Analysis (CCA)~\cite{hardoon2004canonical} or Generative Adversarial Networks (GANs)~\cite{goodfellow2014generative}.
Specifically, methods based on CCA attempt to project samples of different modalities into a common embedding space by maximizing their correlation~\cite{yan2015deep, wang2015deep, wang2016learning, eisenschtat2017linking}, and those based on GANs align samples of different modalities by learning modality-adversarial embeddings~\cite{yin2017adversarial, wang2017adversarial, wang2019learning}.
Unfortunately, these methods cannot consider semantic relations between classes.

This paper shows that the prototype-based embedding learning is fairly effective for cross-modal retrieval. 
Also, unlike the previous work, our method can consider relations between categories, improving performance substantially.


\section{Our Method}
\begin{figure*}[t!]
    \centering
    \includegraphics[width=\textwidth]{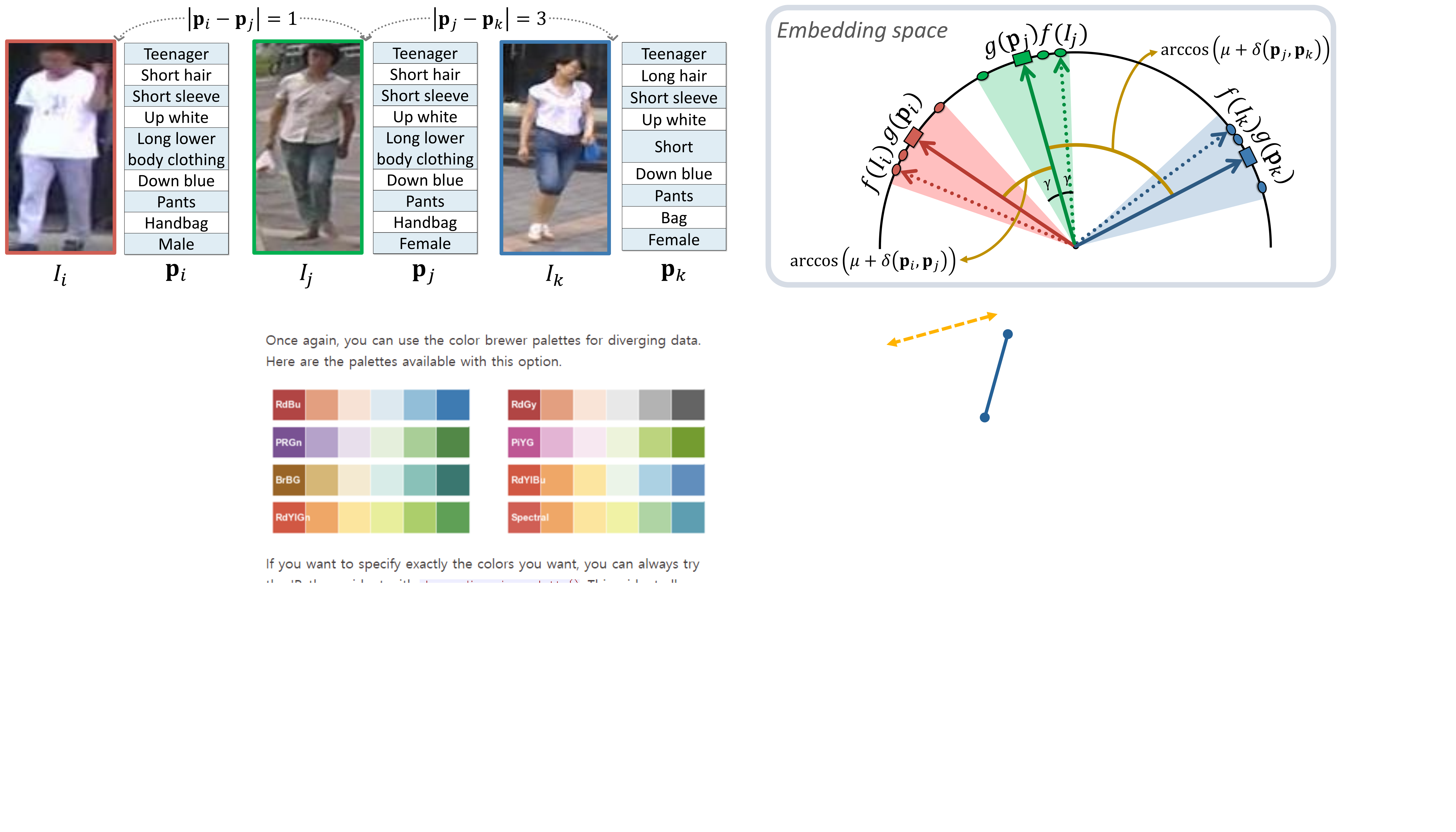}
    \caption{
    A conceptual illustration of the learning objective in Eq.~\eqref{eq:total_loss}. 
    The modality alignment loss pulls images close to their person categories within the margin $\gamma$.
    Meanwhile, ASMR controls margins between person categories according to their semantic affinities.
    }
    \label{fig:Method}
    \vspace{-3mm}
\end{figure*}

In attribute-based person search, a set of attributes, called \emph{person category}, describes traits of people we want to find. 
Given a person category as query, our method conducts person search by finding images that are closest to the query in a joint embedding space of person images and categories.
It learns the embedding space through two encoders, one for images and the other for person categories;
an overview of the architecture is given in Fig.~\ref{fig:architecture}.

The key contribution of our work is the loss function used for training the networks. 
In the embedding space, the loss pulls a person category and its associated images together, and at the same time, pushes apart a pair of person categories by a margin determined adaptively by their semantic dissimilarity.
Our model achieves outstanding performance and converges quickly thanks to the proposed loss, and is computationally efficient due to the straightforward model architecture and retrieval pipeline.

The remainder of this section first describes details of the model architecture and its pretraining, then elaborates the proposed loss function and discusses its advantages.

\subsection{Model Architecture and Its Pretraining}

In the image encoder of our model, a conventional CNN extracts a feature map of input person image, which is in turn transformed to a single feature vector by Global Average Pooling (GAP) and fed to a Multi-Layer Perceptron (MLP) that produces image embedding. 
Meanwhile, the person category encoder is a MLP that takes person category as input and produces person category embedding. 
Outputs of the two encoders are all $\ell_2$ normalized. 

Since a person category is a set of text attributes, it has to be converted in a numerical form to be fed to its encoder. 
To this end, it is given in a form of binary vector, each of whose dimension indicates the presence of corresponding attribute. 
Suppose that attributes are grouped exclusively into a number of \emph{attribute groups}; 
for example, two attributes \sftype{male} and \sftype{female} belong to the same attribute group \sftype{gender}.
As a person can take only one attribute for each attribute group, an attribute is represented by a one-hot vector whose dimension is the same with the number of attributes in its group. 
The binary vector representation of a person category is then obtained by concatenating such one-hot vectors of its all attributes.

Parameters of our model are initialized randomly, except those of the CNN for which we adopt ImageNet pretrained parameters. 
Unfortunately, the weights for ImageNet classification are suboptimal for capturing subtle appearance features of person images.
We thus pretrain the image encoder for attribute classification, an auxiliary task for learning image representation more suitable to person search.\footnote{For the same reason, existing methods also take advantage of the attribute classification by adopting it as an auxiliary task~\cite{dong2019person} or using it for pretraining their models~\cite{cao2020symbiotic}.} 
Specifically, we append a classification head of four Fully Connected (FC) layers on top of the GAP for each attribute group.
Then each classification head is trained together with the backbone CNN for choosing the correct attribute among those in its attribute group through a multi-class classification loss;
we adopt the softmax cross-entropy loss for this purpose.
After the pretraining, a randomly initialized MLP replaces the attribute classification heads.

\subsection{Learning Objective}

The loss for our model consists of two parts.
One of them is a modality alignment loss designed to group embedding vectors of images together around that of their person category for cross-modal alignment.
The other is Adaptive Semantic Margin Regularizer (ASMR) that controls the margin between a pair of embedding vectors of person categories according to their semantic dissimilarity.
The roles of the two components are illustrated in Fig.~\ref{fig:Method}.

Let $f$ and $g$ be the encoders for images and person categories, respectively.
Training data for learning the encoders are provided by a set of images paired with binary vectors indicating their person categories, $\mathcal{D}=\{I_i,~\mathbf{p}_i\}_{i=1}^m$, where $m$ is the number of training images.
In addition, let $\mathcal{G}$ denote the set of embedding vectors of unique person categories in the training set.
Given embedding vectors of images $\mathbf{f}_i := f(I_i)$ and those of person categories $\mathbf{g}_i := g(\mathbf{p}_i)$, the learning objective for our model is a linear combination of the two terms as follows:
\begin{equation}
    \mathcal{L}\big(\{\mathbf{f}_i,~\mathbf{g}_i\}_{i=1}^m\big) = \mathcal{L}_{\textrm{MA}}\big(\{\mathbf{f}_i,~\mathbf{g}_i\}_{i=1}^m\big) + \lambda \ \mathcal{R}\big(\mathcal{G}\big),
    \label{eq:total_loss}
    \vspace{-1mm}
\end{equation}
where $\mathcal{L}_{\textrm{MA}}$ indicates the modality alignment loss, $\mathcal{R}$ means the ASMR, and $\lambda$ is a weight hyper-parameter.
Details of the two components are described below.

\subsubsection{Modality Alignment Loss}

The role of the modality alignment part is to align the two different modalities in a common embedding space.
Considering each person category embedding as a semantic prototype of associated image embeddings, the cross-modal alignment is done by pulling image embeddings close to their semantic prototypes and pushing them apart from irrelevant prototypes.
This idea is formulated as
\begin{equation}
    \begin{split}
        &\mathcal{L}_{\textrm{MA}}\big(\{\mathbf{f}_i,~\mathbf{g}_i\}_{i=1}^m\big) = \\
    &-\frac{1}{m}\sum_{i=1}^m\log \left(\frac{e^{\sigma \cos(a(\mathbf{f}_i, \mathbf{g}_i)+\gamma )}}{e^{ \sigma \cos(a(\mathbf{f}_i, \mathbf{g}_i)+\gamma )} + \displaystyle\sum\limits_{\mathbf{g}_k\in \mathcal{G} \setminus \mathbf{g}_i} e^{\sigma \cos a(\mathbf{f}_i, \mathbf{g}_k)}}\right),
    \end{split}
    \vspace{-1mm}
    \label{eq:loss_metric}
\end{equation}
where $a(\cdot, \cdot)$ means the angle between its two input vectors, $\sigma > 0$ is a scale factor, and $\gamma > 0$ is a margin between image and person category embeddings.
The above form resembles ArcFace loss~\cite{deng2019arcface}, yet different in that the person category embeddings used as prototypes are not parameters but outputs of another encoder $g$ in our loss.
We empirically found that the simple joint embedding architecture trained solely with this loss is as competitive as previous arts; it can be considered as a simple yet solid baseline for attribute-based person search, and ASMR further improves the performance substantially.

\subsubsection{ASMR}
For accurate person search and generalization to unseen person categories, we expect from the learned embedding space that different person categories lie apart from each other clearly and their distances are larger if they are more dissimilar, \ie, sharing less attributes.
However, the modality alignment loss in Eq.~\eqref{eq:loss_metric} alone does not guarantee this quality of embedding space since it ignores semantic relations between them; 
the loss considers person categories simply as independent class labels.
One of failure cases regarding this issue is that different person categories are often located overly close to each other in the learned embedding space when images of these categories exhibit subtle appearance variations; an example is given in Fig.~\ref{fig:tSNE_ASMR}.

To address this issue, we introduce ASMR that explicitly controls the margin between a pair of person categories according to their semantic dissimilarity.
The regularizer is given by
\begin{equation}
    \mathcal{R}(\mathcal{G}) = \frac{1}{\mcomb{|\mathcal{G}|}{2}} \sum_{i=1}^{|\mathcal{G}|-1} \sum_{j=i+1}^{|\mathcal{G}|} \big\{ s(\mathbf{g}_i,\mathbf{g}_j) - \mu - \delta(\mathbf{p}_i, \mathbf{p}_j) \big\}^2,
    \label{eq:Reg}
    \vspace{-1mm}
\end{equation}
where $s(\cdot,\cdot)$ denotes the cosine similarity between the two input vectors and $\mu$ is the mean similarity over all pairs of person categories in the embedding space:
\begin{equation}
    \mu = \frac{1}{\mcomb{|\mathcal{G}|}{2}} \sum_{i=1}^{|\mathcal{G}|-1} \sum_{j=i+1}^{|\mathcal{G}|} s(\mathbf{g}_i,\mathbf{g}_j).
    \vspace{-1mm}
\end{equation}
Also, $\delta(\mathbf{p}_i, \mathbf{p}_j)$ quantifies the semantic similarity of a pair of person categories represented as binary vectors $\mathbf{p}_i$ and $\mathbf{p}_j$, and is formulated as an inverse of weighted Hamming distance:
\begin{equation}
    \delta(\mathbf{p}_i, \mathbf{p}_j) = \text{Sigmoid}\bigg(1 - \sum_{k} w_k |\mathbf{p}_i(k)- \mathbf{p}_j(k) | \bigg).
    \label{eq:delta}
    \vspace{-1mm}
\end{equation}
Regarding its shape, the sigmoid function lets this similarity margin respond more sensitively to pairs of more similar person categories, which in general have to be handled more carefully for accurate person search.\footnote{Person categories sharing more attributes are more likely to be close in the embedding space due to their similar appearances, and to affect accuracy of person search whose goal is to find samples \emph{closest} to query.}
Moreover, the weight parameters $w_k$ are trained together with those of the embedding networks, which enables ASMR to estimate importance of individual attributes and relate person categories in consideration of the importance. 

ASMR enforces $s(\mathbf{g}_i,\mathbf{g}_j)$ to approximate $\mu + \delta(\mathbf{p}_i, \mathbf{p}_j)$ so that the degree of similarity between person categories in the binary vector space is reflected by their similarity in the learned embedding space. 
This behavior of ASMR makes distributions of embedding vectors more discriminative by enlarging the margin between person categories.
Also, we believe that it helps our model avoid being biased to image information and generalize better to unseen person categories by reflecting semantic relations between person categories explicitly in the embedding space.


\section{Experiments}
\begin{table}[t!]
    \centering
    \resizebox{0.49\textwidth}{!}{
    \begin{tabular}{l|c|c|c}
    \hline
        Datasets &  PETA & Market-1501      & PA100K \\ \hline
        \# Attributes            & 65        & 27        & 26   \\
        \# Attributes groups                & 17        & 10        & 15   \\ \hline
        \# Train person category & 1,890     & 508       & 500   \\
        \# Train image           & 12,140    & 12,936    & 80,000   \\ \hline
        \# Test person category  & 200       & 484       & 814   \\
        \# Unseen                & 200         & 315       & 168   \\
        \# Test image            & 1,181     & 16,483    & 10,000   \\\hline
    \end{tabular}}
    \caption{Statistics of three benchmarks.} 
    \vspace{-3mm}
    \label{tab:Statistics}
\end{table}

\begin{table*}[ht]
    \centering
    \resizebox{\textwidth}{!}{
    \begin{tabular}{l|c|c c c|c| c c c |c| c c c|c}
    \hline
         \multirow{2}{*}{Method}& \multirow{2}{*}{Dim}  & \multicolumn{4}{c|}{PETA} & \multicolumn{4}{c|}{Market-1501 Attribute} & \multicolumn{4}{c}{PA100K}  \\  \cline{3-14}
         &  & Rank1 & Rank5 & Rank10 & mAP & Rank1 & Rank5 & Rank10 & mAP  & Rank1 & Rank5 & Rank10 & mAP \\ \hline        
         DeepMAR~\cite{li2015multi} & - & 17.8 & 25.6 & 31.1 & 12.7 & 13.2 & 24.9 & 32.9 & 8.9 & - & - & - & - \\
         DCCAE~\cite{wang2015deep} & - & 14.2 & 22.1 & 30.0 & 14.5 & 8.1 & 24.0 & 34.6 & 9.7 & 21.2 & 39.7 & 48.0 & 15.6 \\
         2WayNet~\cite{eisenschtat2017linking} & - & 23.7 & 38.5 & 41.9 & 15.4 & 11.3 & 24.4 & 31.5 & 7.8 & 19.5 & 26.6 & 34.5 & 10.6 \\
         CMCE~\cite{li2017identity} & - & 31.7 & 39.2 & 48.4 & 26.2 & 35.0 & 51.0 & 56.5 & 22.8 & 25.8 & 34.9 & 45.4 & 13.1 \\
         
         AAIPR~\cite{yin2017adversarial}& 128 & 39.0 & 53.6 & 62.2 & 27.9 & 40.3 & 49.2 & 58.6 & 20.7 & - & - & - & - \\
         AIHM~\cite{dong2019person}& 3K & - & - & - & - & 43.3 & 56.7 & 64.5 & 24.3 & \underline{31.3} & \underline{45.1} & \underline{51.0} & \underline{17.0} \\
         SAL~\cite{cao2020symbiotic}& 128 & \underline{47.0} & \underline{66.5} & \underline{74.0} & \underline{41.2} & \underline{49.0} & \textbf{68.6} & \textbf{77.5} & \underline{29.8} &  - &- & - & - \\
        SAL~\cite{cao2020symbiotic}$^\dagger$& 128 & 39.0 & 61.5 & 70.0 & 37.2 & 44.4 & 65.7 & 72.5 & 29.4 & 22.7 & 36.5 & 41.6 & 15.0 \\ \hline
         Ours& 128 & \textbf{56.5} & \textbf{80.0} & \textbf{83.5} & \textbf{50.2}  & \textbf{49.6} & \underline{64.9} & \underline{72.5}&	\textbf{31.0} & \textbf{31.9} & \textbf{49.1} &\textbf{58.2} & \textbf{20.6}
          \\
         \hline
    \end{tabular}}
    \vspace{0.2mm}
    \caption{
    Quantitative comparison to previous arts. 
    \textsf{\footnotesize{Dim}} indicates embedding dimensions of the methods based on cross-modal embeddings.
    \textbf{Bold} and \underline{underline} denote the best and the second-best, respectively. 
    $\dagger$ indicates results reproduced by the official implementation.
    }
    \label{tab:Results}
    \vspace{-3mm}
\end{table*}

Our method is evaluated and compared to previous work on three public benchmarks for attribute-based person search~\cite{Market1501_attribute, PETA, PA100K}.
We also demonstrate the effect of ASMR by ablation studies and qualitative analysis.

\subsection{Datasets}
We evaluate our method and previous arts on three public datasets, PETA~\cite{PETA}, Market-1501 Attribute~\cite{Market1501_attribute} and PA100K~\cite{PA100K}, which are representative benchmarks for attribute-based person search.
The dataset statistics are summarized in Table~\ref{tab:Statistics}.
Note that the PETA dataset follows the ordinary image retrieval setting where categories of test images are all unseen, while the other two datasets assume a more general search scenario in which both seen and unseen person categories appear in testing.

\subsection{Implementation Details}
\noindent\textbf{Network architecture.}
In the image encoder, the backbone CNN is ResNet-50~\cite{resnet} and the MLP consists of three FC layers. 
On the other hand, the person category encoder is implemented only by a MLP with three FC layers.
Both of the two encoders produce 128-dimensional embedding vectors that are $\ell_2$ normalized. 
More details of the encoders are presented in the supplementary material.

\vspace{1mm}\noindent\textbf{Hyper-parameters.} In every experiment, our model is optimized by SGD with a momentum of 0.9 and a weight decay of $\expnum{5}{4}$ for 10 epochs; each mini-batch consists of 128 images and their person categories. 
The initial learning rate is set to $\expnum{1}{3}$ for the image encoder, and $\expnum{1}{2}$ for the person category encoder and the parameters of the weighted Hamming distance.
Then both learning rates are decayed by a factor of 0.1 at every 5 epochs. 
The other hyper-parameters, $\lambda$ in Eq.~\eqref{eq:total_loss}, and $\sigma$ and $\gamma$ in Eq.~\eqref{eq:loss_metric} are set to (4, 32, 0.1) on PETA, (6, 12, 0.2) on Market-1501 Attribute, and (5, 48, 0.1) on PA100K, respectively. 

\subsection{Quantitative Comparison to Previous Work}

Our model is compared to the three existing methods for attribute-based person search, AAIPR~\cite{yin2017adversarial}, AIHM~\cite{dong2019person}, and SAL~\cite{cao2020symbiotic}.
We also report performance of related models that are not originally proposed for attribute-based person search but have been reproduced for the purpose in literature.
Performance of these methods including ours is summarized in Table~\ref{tab:Results}, where Cumulative Matching Characteristic (CMC) and mean Average Precision (mAP) are adopted as performance metrics following the convention.

\begin{table}[t!]
    \centering
    \resizebox{0.35\textwidth}{!}{
    \begin{tabular}{l|c|c|c}
    \hline
        Method &  PETA & Market-1501& PA100K \\ \hline
        SAL~\cite{cao2020symbiotic} & 202 & 211 & 957   \\
        Ours & 27 & 18 & 110  \\
        \hline
    \end{tabular}}
    \caption{Comparison of training time (min)} 
    \vspace{-3mm}
    \label{tab:Time}
\end{table}

The table shows that our model outperforms all the other methods in terms of Rank1 and mAP metrics.
It clearly surpasses AIHM~\cite{dong2019person}, the state of the art in PA100K, for all available settings. 
This achievement is remarkable since our method is more efficient than AIHM; it works with embedding vectors of a substantially smaller dimension, and unlike AIHM, it does not require any extra network for retrieval.
Moreover, our method outpaces SAL~\cite{cao2020symbiotic}, the state of the art in PETA and Market-1501 Attribute, for almost all settings. Especially, it outperforms SAL on the PETA dataset by a large margin, 9.5\% in Rank1 and 9.0\% in mAP.
On the Market-1501 Attribute dataset, it is more accurate than SAL in terms of Rank1 and mAP, although its records in Rank5 and Rank10 are slightly below those of SAL.

The key to this success of our method is two-fold. 
The first is MA loss.
Since the loss compares each image embedding with those of all person categories in the dataset, it enables to learn more discriminative embedding space more efficiently.
Meanwhile, the loss of AIHM considers images and person categories within a mini-batch only.
Another cause is the person category encoder, which encodes person categories in an attribute-aware manner so that the embedding space reflects their semantic relations.
On the other hand, SAL represents person categories as independent network parameters.
The last yet most vital cause is ASMR, whose efficacy is validated in Sec.~\ref{sec:Effect}.

The reasons for the small improvement on the Market-1501 Attribute and PA100K datasets are as follows.
Compared to the PETA dataset, these datasets assume a more challenging search scenario in which both seen and unseen person categories appear in testing. 
Further, in the Market-1501 dataset, incorrect attribute labels bind the performance; this happens because the labels are annotated not per image but per video, \eg, a man labeled with ``jacket'' may take off his jacket in the middle of video.

\begin{figure}[t!]
    \centering
    \includegraphics[width=.48\textwidth]{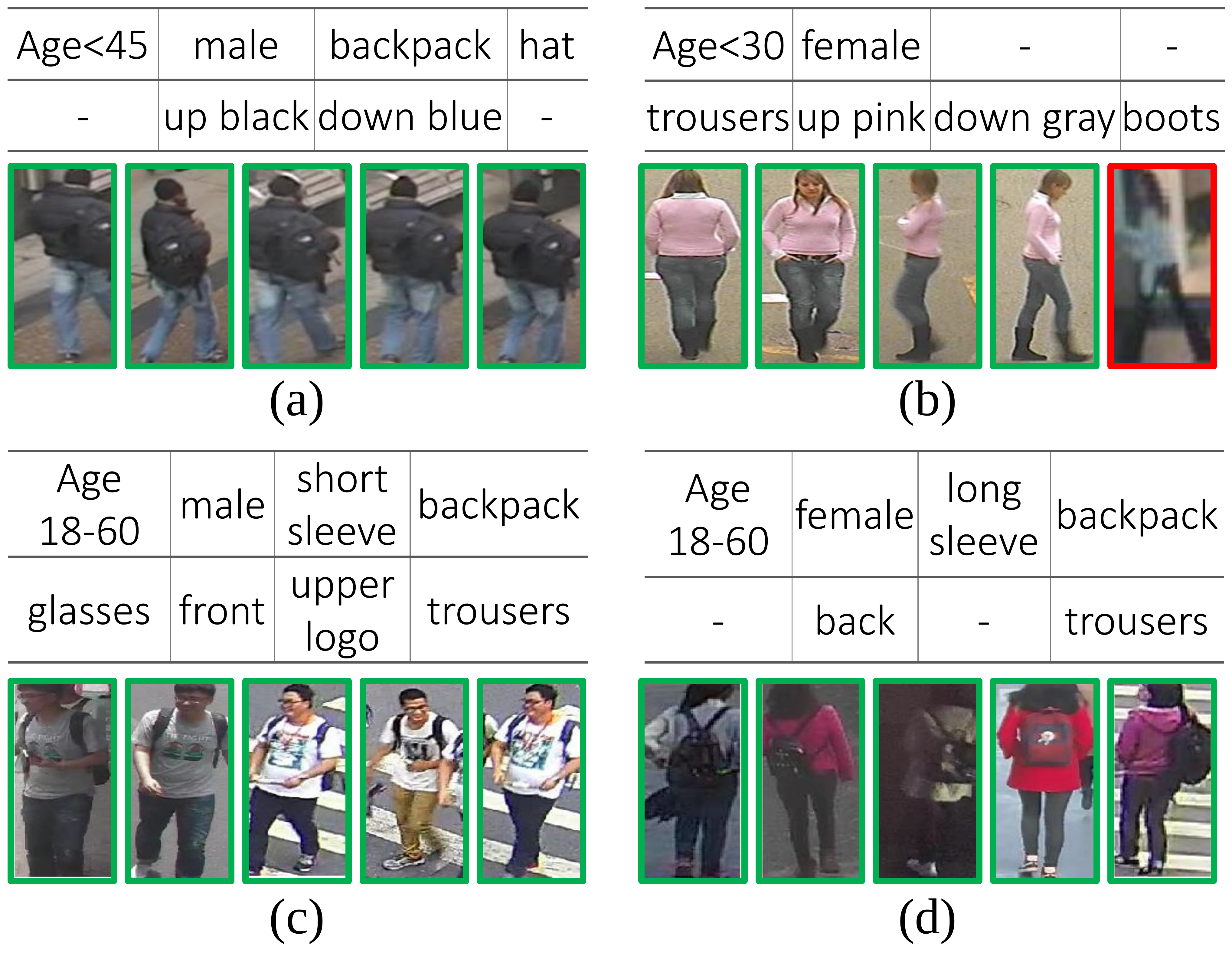}
    \caption{
    Top 5 retrieval results of our method on (a, b) the PETA and (c, d) PA100K datasets.
    Images are sorted from left to right according to their ranks.
    Green and red boxes indicate true and false matches, respectively.
    Queries are given as tables, where blanks indicate attributes that do not exist in the query.
    }
    \label{fig:Qual_PETA}
    \vspace{-3mm}
\end{figure}

In addition, compared to SAL, our model is significantly more efficient in training. 
SAL requires a large training time for convergence due to its adversarial learning strategy.
In contrast, our method is trained simply by supervised learning with the loss function in Eq.~\eqref{eq:total_loss}.
In consequence, ours using a single GPU converges more than 7.5 times faster than SAL using two GPUs as shown in Table~\ref{tab:Time}.

\begin{figure}[t!]
    \centering
    \includegraphics[width=.48\textwidth]{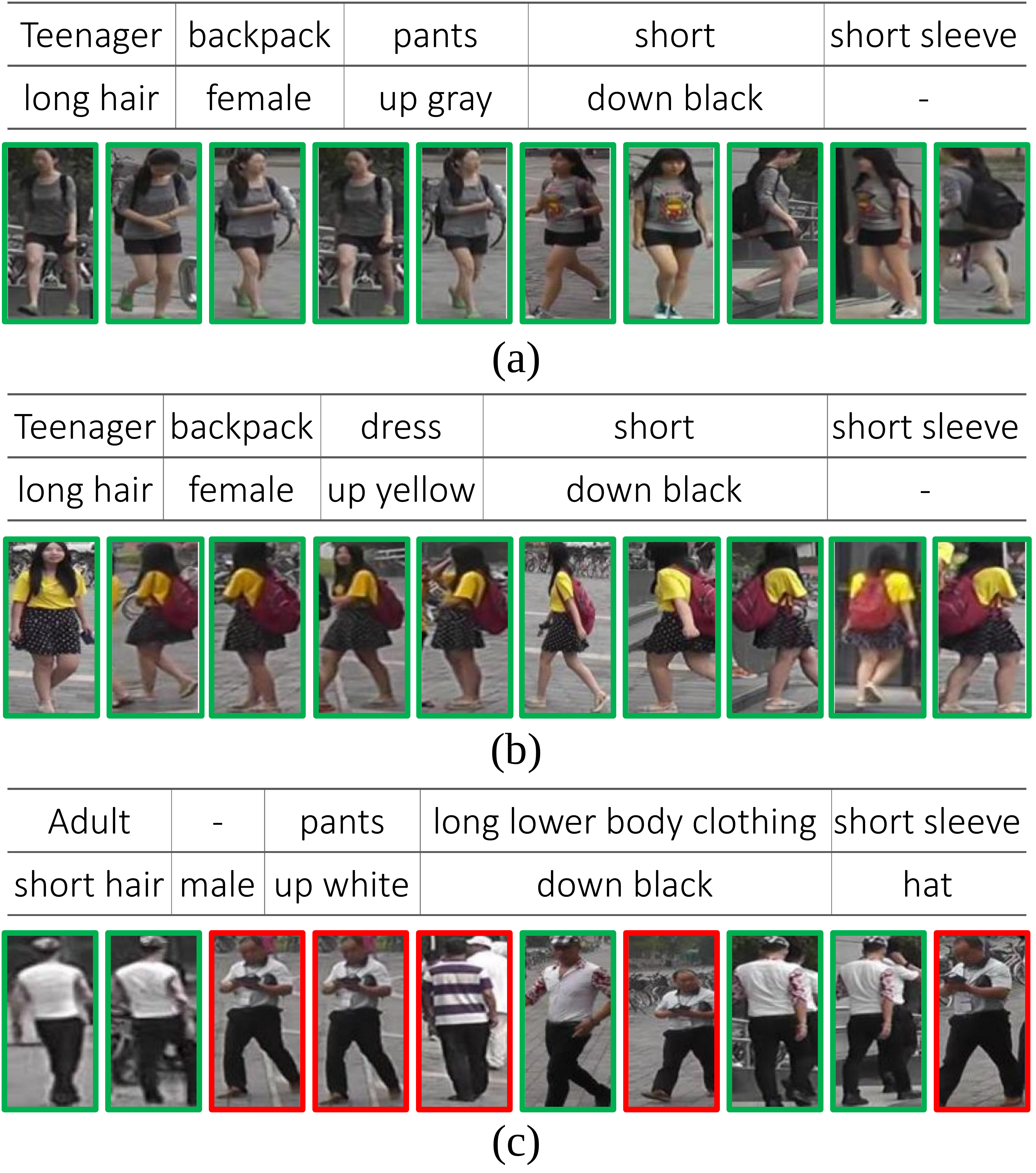}
    \caption{
    Top 10 retrieval results of our method on the Market-1501 Attribute dataset.
    Images are sorted from left to right according to their ranks.
    Green and red boxes indicate true and false matches, respectively.
    Queries are presented above their retrieved images; blanks indicate attributes that do not exist in the query.
    }
    \label{fig:Qual_Market}
    \vspace{-1mm}
\end{figure}

\subsection{Qualitative Analysis}
Qualitative results of the proposed method
are given in Fig.~\ref{fig:Qual_PETA} and Fig.~\ref{fig:Qual_Market}. 
All the presented results demonstrate that our method is insensitive to severe variations in human and camera poses.
Moreover, individual examples show that our method is robust against changes in image resolution (Fig.~\ref{fig:Qual_PETA}(b,c,d), Fig.~\ref{fig:Qual_Market}(a,b,c)), illumination (Fig.~\ref{fig:Qual_PETA}(c,d), Fig.~\ref{fig:Qual_Market}(a,b,c)), and partial occlusions (Fig.~\ref{fig:Qual_PETA}(a,c,d)).
It is also demonstrated that the proposed method is able to capture fine details of images for precise retrieval; examples include \sftype{backpack} in Fig.~\ref{fig:Qual_PETA}(a,c,d) and in Fig.~\ref{fig:Qual_Market}(a,b), \sftype{hat} in Fig.~\ref{fig:Qual_PETA}(a) and Fig.~\ref{fig:Qual_Market}(c) and \sftype{glasses} and \sftype{clothing pattern} in Fig.~\ref{fig:Qual_PETA}(c).
More qualitative results can be found in the supplementary material.

\begin{table}[t!]
    \footnotesize
    \centering
    \begin{tabular}{cl|c|c|c}
        \hline
        \multicolumn{2}{l|}{Method}  & PETA & Market & PA100K \\\hline        
        \multirow{2}{*}{(a)} & \multicolumn{1}{|l|}{Baseline} & 46.5 & 30.4 & 26.0 \\
         & \multicolumn{1}{|l|}{Baseline + pretraining} & 48.5 & 44.8 & 28.9 \\ 
         \hline
        
        \multirow{4}{*}{(b)} & \multicolumn{1}{|l|}{$\mathcal{L}_\textrm{MA}\rightarrow$ Proxy Anchor~\cite{Kim_2020_CVPR}}  & 48.0 & 41.1 & 27.4 \\
        & \multicolumn{1}{|l|}{$\mathcal{L}_\textrm{MA}\rightarrow$ Proxy NCA~\cite{movshovitz2017no}} & 52.0 & 43.8 & 29.7 \\
        & \multicolumn{1}{|l|}{${\mathcal{L}_\textrm{MA}\rightarrow}$ CosFace~\cite{wang2018cosface}} & {50.5} & {45.3} & {24.9} \\
        & \multicolumn{1}{|l|}{${\mathcal{L}_\textrm{MA}\rightarrow}$ {SphereFace~\cite{Liu2017CVPR}}} & {52.5} & {45.0} & {23.8} \\\hline

        & Ours & \textbf{56.5} & \textbf{49.6} & \textbf{31.9}\\ \hline     
    \end{tabular}
    
    \vspace{1.0mm}
    \caption{
    Performance in Rank@1 of ours and its variants on the PETA, Market-1501 Attribute, and PA100K datasets.
    }
    \label{tab:Ablation}
    \vspace{-3mm}
\end{table}

\subsection{Ablation Studies}

\noindent \textbf{Effects of pretraining and ASMR.}
We quantify the effects of our pretraining strategy and ASMR by evaluating two reduced versions of our method with and without them.
To this end, we first define a baseline as the model with the same architecture as ours yet trained only with $\mathcal{L}_{\textrm{MA}}$ in Eq.~\eqref{eq:loss_metric};
the other variant is obtained by adding the pretraining to the baseline.
The results in Table~\ref{tab:Ablation}(a) suggest that the contribution of ASMR is significant and the pretraining also helps to some extent.
In detail, ASMR contributes to the performance, enhancing Rank1 by 8.0\% on PETA, 4.8\% on Market-1501 Attribute, and 3.0\% on PA100K, respectively.
These results suggest that ASMR makes the learned embedding space more discriminative and better generalized to unseen categories. 
Also, the pretraining improves Rank1 by 2.0\% on PETA, 14.4\% on Market-1501 Attribute, and 2.9\% on PA100K, respectively, which clearly validates its effectiveness.
Further, the table shows that the simple baseline is already comparable to the state of the art; 
we believe that it is a solid and unexplored baseline that future work has to consider.
Finally, we again emphasize that state-of-the-art methods~\cite{dong2019person,cao2020symbiotic} also take advantage of attribute classification, thus the comparison in Table~\ref{tab:Results} is equitable.

\vspace{1mm} \noindent\textbf{Comparison to other embedding losses}.
To demonstrate superiority of our modality alignment loss $\mathcal{L}_\textrm{MA}$, we evaluate variants of our method that replace $\mathcal{L}_\textrm{MA}$ with Proxy Anchor~\cite{Kim_2020_CVPR}, Proxy NCA~\cite{movshovitz2017no}, CosFace~\cite{wang2018cosface} and SphereFace~\cite{Liu2017CVPR}, representative embedding losses using prototypes.
Table~\ref{tab:Ablation}(b) shows that our method using $\mathcal{L}_\textrm{MA}$ largely outperformed the two variants, which indicates the advantage of $\mathcal{L}_\textrm{MA}$.

\begin{table}[t!]
    \small
    \centering
    \begin{tabular}{l|c|c|c}
        \hline
        Method  & PETA & Market-1501 & PA100K \\\hline        
        w/o $\delta(\mathbf{p}_i, \mathbf{p}_j)$ & 52.0 & 46.1 & 30.3 \\
        Uniform $w_k$ & 52.5 & 46.5 & 29.8 \\
        $\ell_2$ normalized $w_k$ & 52.0 & 46.3 & 30.1 \\ \hline
        Ours & \textbf{56.5} & \textbf{49.6} & \textbf{31.9}\\ \hline     
    \end{tabular}
    \vspace{1mm}
    \caption{
    Comparison of ASMR and its variants in Rank@1 of the search resutls on the three datasets.
    }
    \label{tab:Ablation_ASMR}
\end{table}

\begin{figure}[t!]
    \centering
    \includegraphics[width=.48\textwidth]{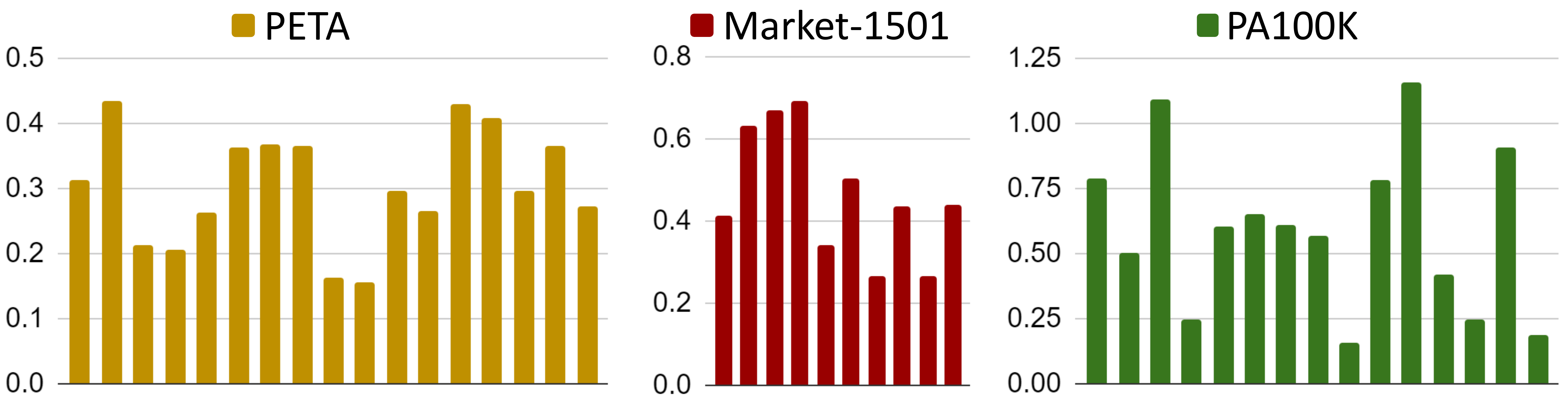}
    \caption{Visualization of $w_k$ learned in our method on the three datasets, 
    where each value corresponds each attribute.
    }
    \label{fig:lambda}
    \vspace{-2mm}
\end{figure}

\begin{figure*}[t!]
    \centering
    \includegraphics[width=\textwidth]{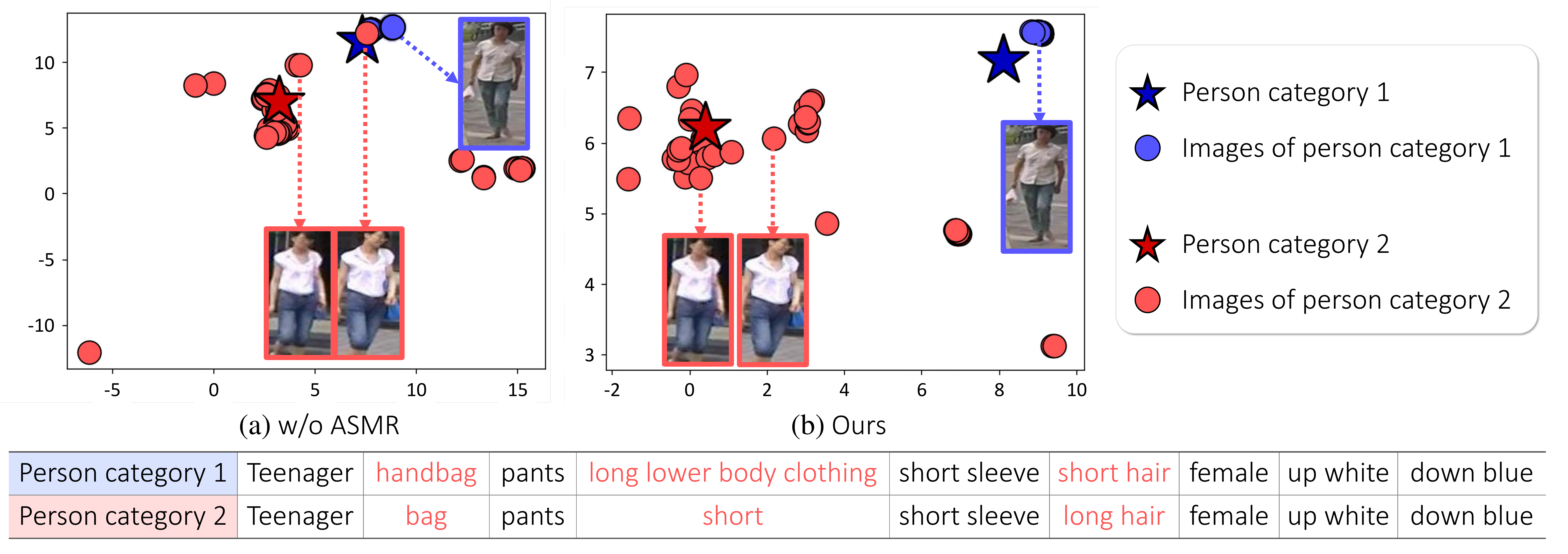}
    \caption{
    $t$-SNE visualization of a part of the joint embedding space learned for the Market-1501 Attribute dataset. Stars and circles indicate embedding vectors of person categories and their associated images, respectively, and their colors mean their person categories. The person categories are elaborated below, where attributes that are different between the two categories are colored in red. 
    }
    \label{fig:tSNE_ASMR}
    \vspace{-3mm}
\end{figure*}

\begin{figure}[t!]
    \centering
    \includegraphics[width=.49\textwidth]{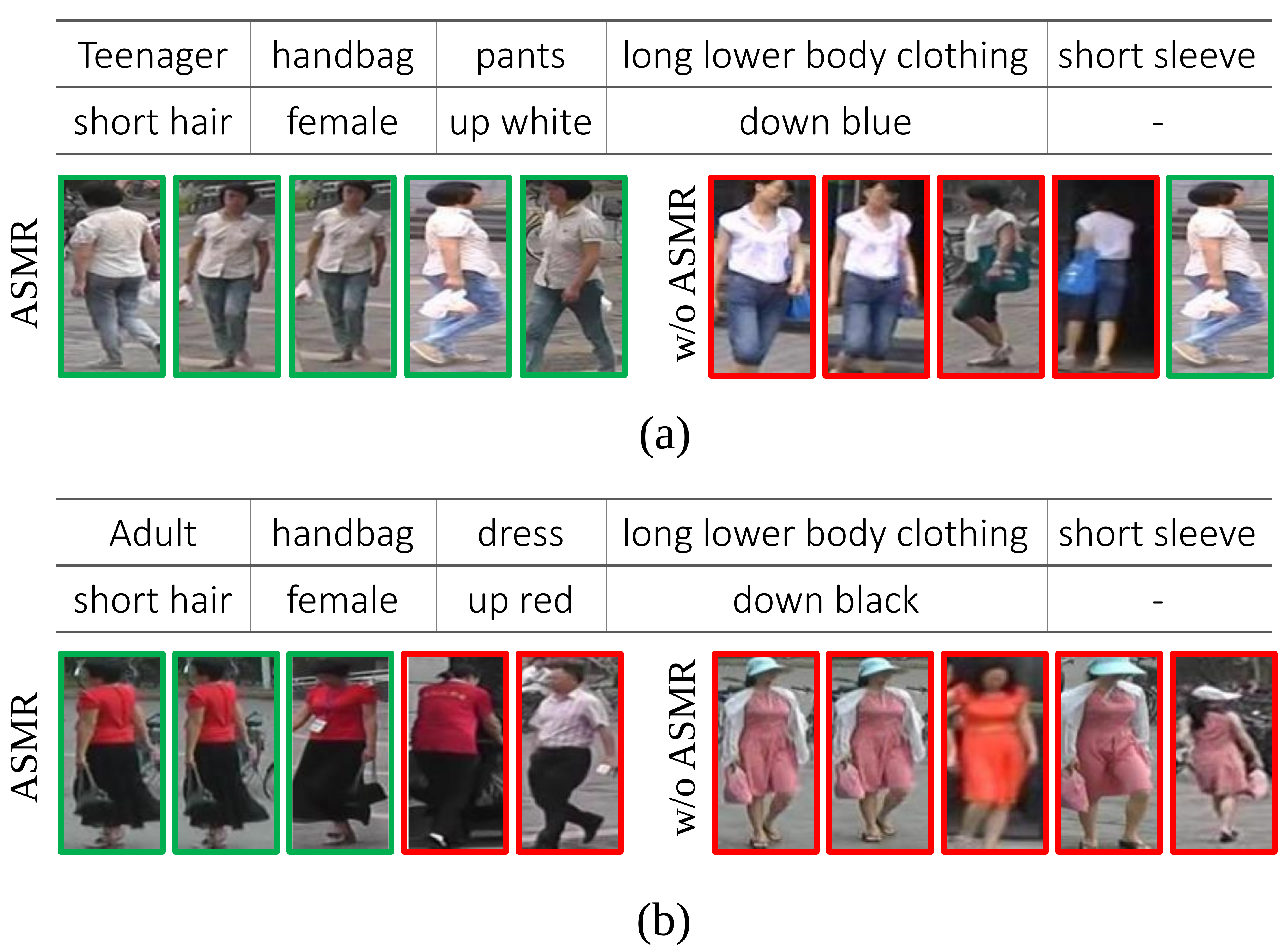}
    \caption{
    Top 5 retrieval results of our method and its variant without ASMR on the Market-1501 Attribute dataset.
    Images are sorted from left to right according to their ranks.
    Green and red boxes indicate true and false matches, respectively. 
    }
    \label{fig:Qual_ASMR}
    \vspace{-4mm}
\end{figure}

\subsection{In-depth Analysis on ASMR}
\label{sec:Effect}
The effect of each design points of ASMR are verified by experiments, whose results are summarized in Table~\ref{tab:Ablation_ASMR}.
First of all, the large gap between ours and its variant without $\delta$ demonstrates the significant contribution of $\delta$ to the performance. 
Note that ASMR without $\delta$ becomes analogous to the diversity regularizer in~\cite{hayat2019gaussian}, and forces person category embeddings to be uniformly distributed. 
This suggests that ASMR does not blindly enlarge between-category margins but controls them with consideration to semantic affinities between categories, which is vital for the outstanding performance of our work.

The role of learnable weights $w_k$ in $\delta$ is also investigated.
We observed that the performance drops when the weights are fixed by a single value (\ie, uniform $w_k$), which suggests that the learned weights well capture the unequal importance of attributes.
We also found that imposing $\ell_2$ normalization to the weights does not useful, rather damages performance; Fig.~\ref{fig:lambda} shows that our method learns non-uniform and positive weights with no such a constraint.

Finally, we present more detailed qualitative analysis on the effect of ASMR to explain how it works and to validate its contribution. 
Fig.~\ref{fig:tSNE_ASMR} compares joint embedding spaces learned by our model and its reduced version without ASMR. 
We adopt $t$-SNE~\cite{MaatenNov2008} to visualize their embedding spaces, and focus only on two particular person categories sharing many of their attributes for a clear analysis.
As shown in Fig.~\ref{fig:tSNE_ASMR}(a), some images of person category 2, whose appearances are quite similar to those of person category 1, are located overly close to person category 1 in the embedding space learned without the regularizer; such images will lead to failures in person search. 
This happens since the model is biased towards the image modality immoderately if no constraint is imposed for person category embeddings. 
In contrast, Fig.~\ref{fig:tSNE_ASMR}(b) shows that our final model with the regularizer enlarges the margin between the two categories according to their semantic dissimilarity so that they are well discriminated in the embedding space.

The effectiveness of ASMR is further validated by comparing retrieval results of the models with and without the regularizer in Fig.~\ref{fig:Qual_ASMR}. 
The results suggest that the model without the regularizer often fails when images of different person categories are overly similar as in Fig.~\ref{fig:Qual_ASMR}(a) and/or some attributes of query are about fine details of images like \sftype{hat} and \sftype{age} in Fig.~\ref{fig:Qual_ASMR}(b).
Our method with ASMR handles these issues effectively thanks to the improved discriminability by ASMR.


\section{Conclusion}

We have presented an efficient and effective framework for attribute-based person search.
The main contribution of our work is a novel loss function based on ASMR for learning cross-modal embeddings: It aligns a person category and associated images in a common embedding space, and at the same time, arranges person categories according to their semantic affinities in the space. 
We demonstrated by experiments that the proposed loss allows a simple embedding model to achieve state-of-the-art performance.
Considering its brevity and outstanding performance, our work will be a solid baseline for attribute-based person search.

\vspace{3mm}
{
\noindent \textbf{Acknowledgement:} 
This work was supported by 
the NRF grant, 
the IITP grant, 
and R\&D program for Advanced Integrated-intelligence for IDentification, 
funded by Ministry of Science and ICT, Korea
(No.2019-0-01906 Artificial Intelligence Graduate School Program--POSTECH,
 NRF-2021R1A2C3012728--30\%, 
 NRF-2018R1A5A1060031--20\%, 
 NRF-2018M3E3A1057306--30\%, 
 IITP-2020-0-00842--20\%). 
}

\newpage
{\small
\bibliographystyle{ieee_fullname}
\bibliography{output}
}
\newpage

\renewcommand*{\thefootnote}{\arabic{footnote}}
\renewcommand\thesection{\Alph{section}}
\setcounter{section}{0}
\section{Appendix}

\subsection{Architecture Details}

This section describes details of our model architecture that consists of two encoders, \emph{image encoder} and \emph{person category encoder}.
Configurations of the two encoders are elaborated in Table~\ref{tab:Encoder}, where $d_{pc}$ denotes the dimension of a person category vector, a binary vector obtained by concatenating one-hot vectors of its all attributes. 
Specifically, $d_{pc}$ is 105, 30, and 26 for the PETA~\cite{PETA}, Market-1501 Attribute~\cite{Market1501_attribute}, and PA100K~\cite{PA100K} datasets, respectively.

\begin{table}[ht!]
    \centering
    \begin{tabular}{c| c| c| c}
        \hline
        \multicolumn{2}{c|}{Image encoder} & \multicolumn{2}{c}{Person category encoder} \\ \hline\hline
        Structure & Size  & Structure & Size   \\ \hline
        ResNet-50 & \multirow{2}{*}{2048} & &\\
        + GAP & & & \\ \hline
        \multirow{2}{*}{$\text{FC}_1$} & 2048 $\times$ 512 & \multirow{2}{*}{$\text{FC}_1$} & $d_{pc}$ $\times$ 512 \\
        & ReLU & & ReLU \\ \hline
        \multirow{2}{*}{$\text{FC}_2$} & 512 $\times$ 128 & \multirow{2}{*}{$\text{FC}_2$} & 512 $\times$ 128 \\
        & ReLU & & ReLU \\ \hline
        $\text{FC}_3$ & 128 $\times$ 128 & $\text{FC}_3$ & 128 $\times$ 128 \\ \hline
        \end{tabular}%
    \vspace{0.1mm}
    \caption{Details of the two encoders.}\label{tab:Encoder}
    \vspace{-2mm}
\end{table}
\subsection{Pretraining of Image Encoder}
\label{sec:pretrain}
Before training of our model, the image encoder is pretrained by multiple attribute classifiers to make the image representation more suitable to person search.  
As shown in Figure~\ref{fig:Pretraining}, we append a classifier head with four Fully Connected (FC) layers on top of Global Average Pooling (GAP) for each attribute group.
Each classifier is learned to choose the correct attribute among those in each attribute group. 
It consequently improves the representation power of the image encoder backbone.
Specifically, the image encoder is pretrained by the softmax cross-entropy loss per classifier:
\begin{equation}
    \mathcal{L}_{\text{cls}} = -\frac{1}{m}\sum_{i=1}^m\sum_{j=1}^{n}\log\left(p(I_i,y_{ij})\right),
\end{equation} 
where $p(I_i,y_{ij})$ means the predicted probability of the $i$-th training image $I_i$ for its groundtruth attribute of the $j$-th attribute group $y_{ij}$ while $m$ and $n$ denote the number of training images and that of attribute groups, respectively. 
The classification loss $\mathcal{L}_{cls}$ is applied at the pretraining stage only, and the pretrained CNN is utilized as the backbone of the image encoder at the next stage.
\begin{figure}[t!]
    \centering
    \includegraphics[width=0.49\textwidth]{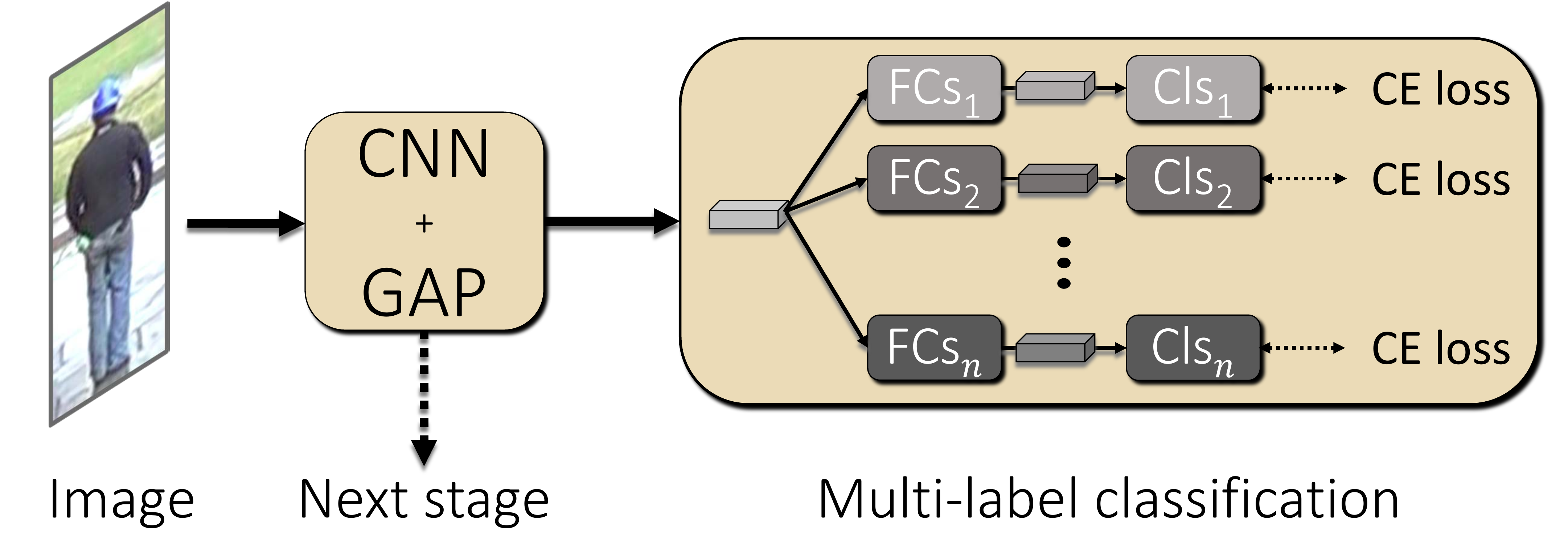}
    \caption{Pretraining of the image encoder. \sftype{Cls} and \sftype{CE loss} denote the attribute classifiers and Softmax Cross Entropy loss, respectively. 
    After pretraining, the multi-label classification part is removed and only the CNN backbone is used for the next stage.}
    \label{fig:Pretraining}
    \vspace{-3mm}
\end{figure}

\subsection{Details about Attribute Groups}
A person category is represented as a binary vector, which consists of exclusive attribute groups such as \sftype{Gender}, \sftype{Age} and \sftype{Accessory}.
We define and utilize the attribute groups for both pretraining of the image encoder and calculating the weighted Hamming distance.
The attribute groups of each dataset we adopt are enumerated in Table~\ref{tab:dataset}.

\begin{figure}[t!]
  \centering
    \includegraphics[width=0.4\textwidth]{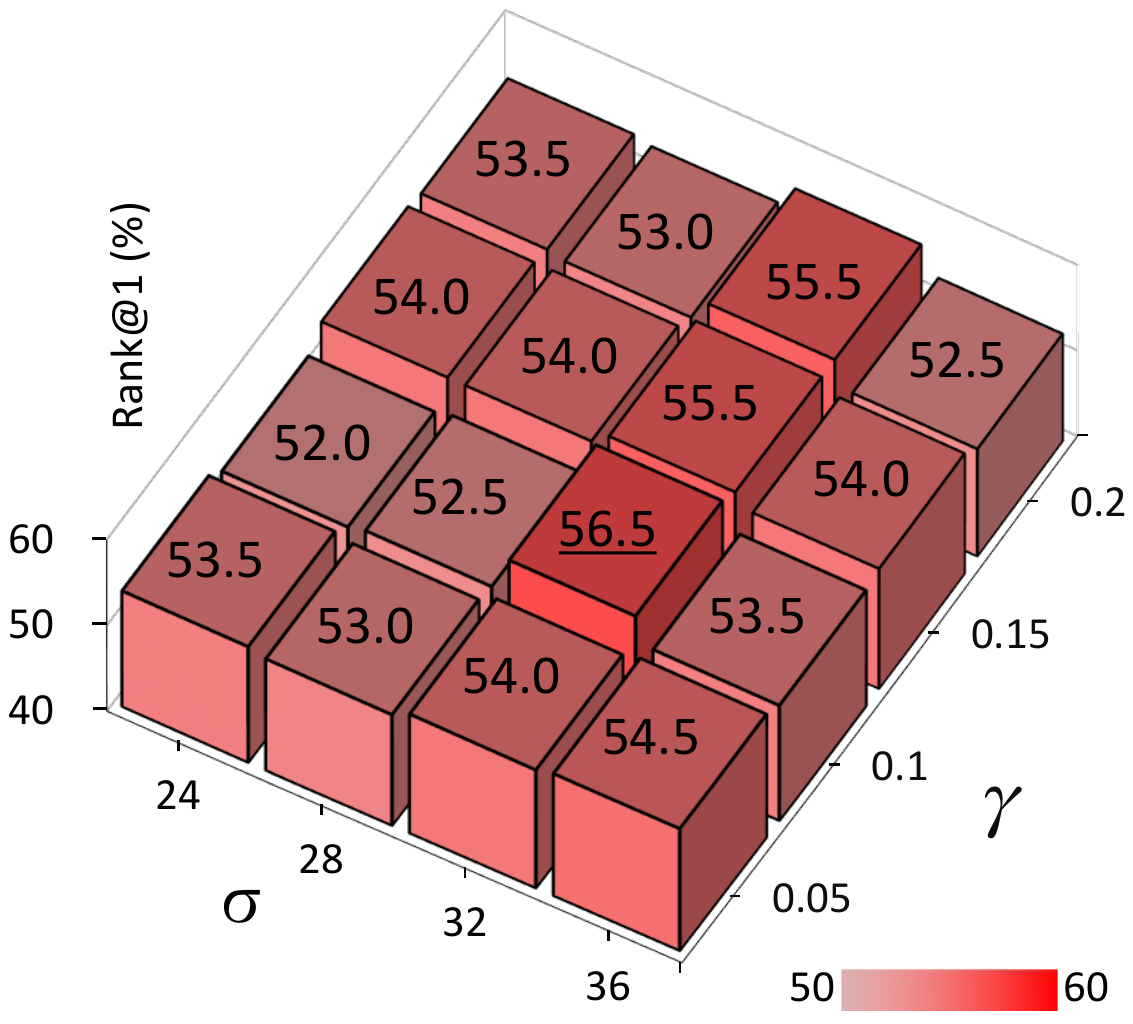}
    \caption{Rank-1 versus $\sigma$ and $\gamma$ on the PETA dataset.}
    \label{fig:ablation_hyper}
  \vspace{-3mm}
\end{figure}

\begin{table*}[ht!]
    \centering
    
    \resizebox{\textwidth}{!}{
    \begin{tabular}{l| c}
    \hline
        Dataset & Attribute group \\ \hline
        \multirow{3}{*}{PETA~\cite{PETA}} & Age, Carrying, Upper body casual, Lower body casual, Accessory, Footwear,\\ 
        & Kind of upper body, Sleeve, Kind of lower body, Texture of upper body, Texture of lower body,\\
        & Gender, Hair length, Color of upper body, Color of lower body, Color of footwear, Color of hair \\\hline
        \multirow{2}{*}{Market-1501 Attribute~\cite{Market1501_attribute}} & Age, Bag, Color of lower body clothing, Color of upper body clothing, Type of lower body clothing, \\
        &Length of lower body clothing, Sleeve length, Hair length, Hat, Gender \\ \hline 
        
        \multirow{2}{*}{PA100K~\cite{PA100K}} & Age, Gender, Viewpoint, HandBag, ShoulderBag, Backpack, HoldObjectsInFront, Hat, Glasses\\
        &  Length of Sleeve, Patterns of upper body, Patterns of lower body, Coat, Kind of lower body, Boots\\

        \hline
    \end{tabular}}
    \vspace{0.1mm}
    \caption{Lists of attribute groups in the three benchmark datasets for attribute-based person search.
    }
    \vspace{-2mm}
    \label{tab:dataset}
\end{table*}

\begin{figure*}[t!]
  \centering
    \includegraphics[width=\textwidth]{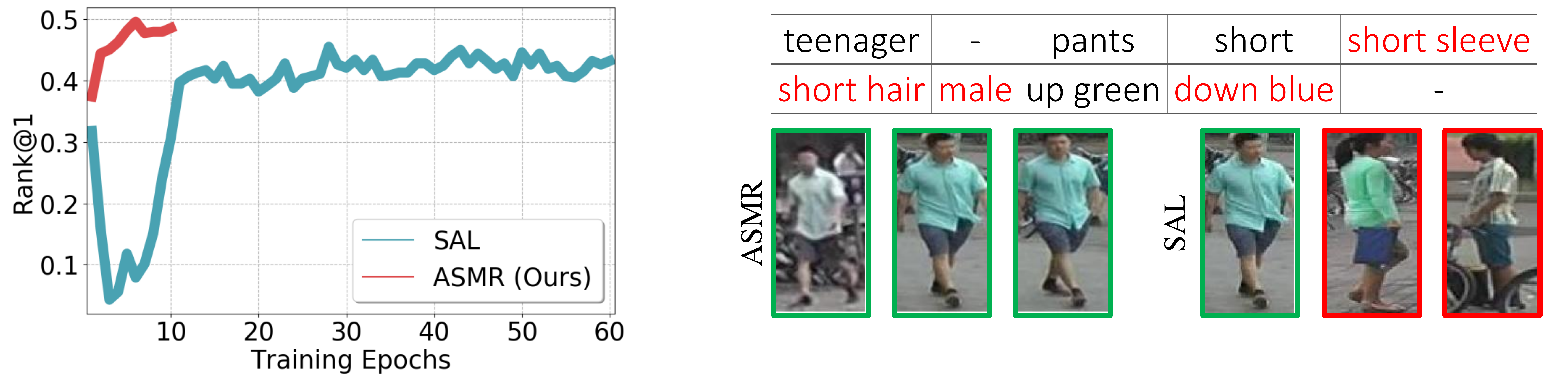}
    \caption{Further comparisons between ours and SAL.
    ({\it left}) Rank-1 versus training epoch on the Market-1501 dataset. ({\it right}) Top 3 retrieval results on the Market-1501 dataset.}
    \label{fig:Training_epoch}
  \vspace{-3mm}
\end{figure*}

\subsection{Effect of a Hyper-parameter}

To investigate the effect of ASMR, an ablation study is conducted by varying the value of $\lambda$, the importance weight for ASMR in Eq.~(1) of the main paper, on the three datasets. 
As shown in Table~\ref{tab:lambda_2}, ASMR improves performance on all the three datasets when $\lambda\geq4$; these results suggest that ASMR is effective regardless of datasets if its importance weight is sufficiently large.
We also stress that, as we tune $\lambda$ for each dataset to obtain optimal scores, the state of the art (\ie, SAL~\cite{cao2020symbiotic}) also tune multiple hyper-parameters (\eg, learning rates for the attribute and image encoders) differently for different datasets to achieve their final scores.
We further conduct an ablation study by varying $\sigma$ and $\gamma$, the scale and margin for MA loss in Eq.~(2) of the paper, on the PETA dataset.
Fig.~\ref{fig:ablation_hyper} demonstrates that our method consistently outperforms state of the art even with diverse hyper-parameter setting.

\begin{table}[t!]
    \footnotesize
    \centering
    \scalebox{0.95}{
    \begin{tabular}{c|cccccccc}
        \hline
        $\lambda$ & 0 & 1 & 2 & 3 & 4 & 5 & 6 & 7 \\ \hline
        PETA & 48.5 & 49.5 & 50.5 & 50.5 & 56.5 & 53.0 & 51.5 & 52.0 \\ \hline
        Market& 44.8 & 44.4 & 44.8 & 45.0 & 46.3 & 47.7 & 49.6 & 45.3 \\ \hline
        PA100K & 28.9 & 28.2 & 27.9 & 28.9 & 30.7 & 31.9 & 30.2 & 29.0 \\ \hline 
    \end{tabular}
    }
    \vspace{0.5mm}
    \caption{
    Performance in Rank@1 versus $\lambda$ on the three datasets.
    }
    \label{tab:lambda_2}
    \vspace{-2mm}
\end{table}

\subsection{More Comparison to SAL}
Our method outperforms SAL in terms of Rank-1 and mAP, yet worse in terms of Rank-5 and Rank-10.
This implies that our retrieval results are more {\it precise} (Rank-1), and are less sensitive to the threshold of CMC (mAP).
In addition, we reproduced SAL~\cite{cao2020symbiotic} through its official implementation\footnote{\href{https://github.com/ycao5602/SAL}{https://github.com/ycao5602/SAL}} to compare our method with SAL in terms of training complexity and performance on the PA100K dataset and qualitative comparison.
As shown in Table~\ref{tab:Reproduce}, the proposed method outperforms both of SAL and its reproduced version (SAL$^\dagger$) on all the three datasets in Rank1 and mAP.
Moreover, we would stress that SAL was not stable in training and the records of SAL reported in the paper were not well reproduced; 
we suspect that the main reason for this failure would be the complicated training procedure of SAL based on adversarial learning.
Lastly, Fig.~\ref{fig:Training_epoch} shows the further comparisons between ours and SAL$^\dagger$ in terms of the training convergence ({\it left}) and the qualitative comparisons ({\it right}), on the  Market-1501 dataset.

\subsection{Failure Cases}
Even though ASMR considers semantic dissimilarity between person categories, it sometimes fails then query attributes are not visually well-distinguishable.
Fig.~\ref{fig:Failure} show examples of such failures due to subtle appearance differences between "adult" and "teenager" or between "bag" and "handbag".

\begin{table}[t!]
    \centering
    \resizebox{0.47\textwidth}{!}{
    \begin{tabular}{l|c|c|c|c|c|c}
    
    \hline
        \multirow{2}{*}{Method} & \multicolumn{2}{c|}{PETA} & \multicolumn{2}{c|}{Market-1501 Attribute} & \multicolumn{2}{c}{PA100K} \\ \cline{2-7}
        & Rank1 & mAP & \quad Rank1~\quad & mAP & Rank1 & mAP \\ \hline
        
        SAL & 47.0 & 41.2 & 49.0 & 29.8 & - & -  \\
        SAL$^\dagger$ & 39.0 & 37.2 & 44.4 & 29.4 & 22.7 & 15.0 \\ \hline
        Ours & \textbf{56.5} & \textbf{50.2} & \textbf{49.6} & \textbf{31.0} & \textbf{31.9} & \textbf{20.6} \\
        \hline
    \end{tabular}}
    \vspace{2mm}
    \caption{Comparison of SAL, its reproduction by the official implementation ($\dagger$), and our method on the three public datasets.} 
    \vspace{-3mm}
    \label{tab:Reproduce}
\end{table}


\subsection{Qualitative Results}
More qualitative results of our method on the three public datasets are presented in Fig.~\ref{fig:Supple_Qual_PETA},~\ref{fig:Supple_Qual_PA100K}, and~\ref{fig:Supple_Qual_Market}.
Results on the PETA and the Market-1501 Attribute datasets overall demonstrate that our method is insensitive to pose variations. 
Also, our method learns body pose variations on the PA100K dataset with viewpoint labels.
In detail, individual results show that our method is robust against changes in image resolution (Fig.~\ref{fig:Supple_Qual_PETA}(a,d-j), Fig.~\ref{fig:Supple_Qual_PA100K}(b,c,f), Fig.~\ref{fig:Supple_Qual_Market}(e-i)), illumination (Fig.~\ref{fig:Supple_Qual_PETA}(b,f-m), Fig.~\ref{fig:Supple_Qual_PA100K}(b,c), Fig.~\ref{fig:Supple_Qual_Market}(a-c)), and partial occlusions (Fig.~\ref{fig:Supple_Qual_PETA}(b,c,l), Fig.~\ref{fig:Supple_Qual_Market}(l)).
Even though some attributes are associated with tiny details of images such as \sftype{hat} (Fig.~\ref{fig:Supple_Qual_PETA}(n), Fig.~\ref{fig:Supple_Qual_Market}(i,j)), \sftype{bag} (Fig.~\ref{fig:Supple_Qual_PETA}(b-e,m-o), Fig.~\ref{fig:Supple_Qual_PA100K}(e-k), Fig.~\ref{fig:Supple_Qual_Market}(e-k)), \sftype{footwear} (Fig.~\ref{fig:Supple_Qual_PETA}(d-o)), \sftype{accessory} (Fig.~\ref{fig:Supple_Qual_PETA}(e), Fig.~\ref{fig:Supple_Qual_PA100K}(c,d)), and \sftype{clothes patterns} (Fig.~\ref{fig:Supple_Qual_PA100K}(a-f,
o)), our method well captures such fine details and retrieves images accurately.
However, our method sometimes fails when query attributes are not visually well-distinguishable (Fig.~\ref{fig:Supple_Qual_PETA}(b,g,j,o), Fig.~\ref{fig:Supple_Qual_PA100K}(f,h,o), Fig.~\ref{fig:Supple_Qual_Market}(k,l)).

\begin{figure*}[t!]
    \centering
    \includegraphics[width=0.9\textwidth]{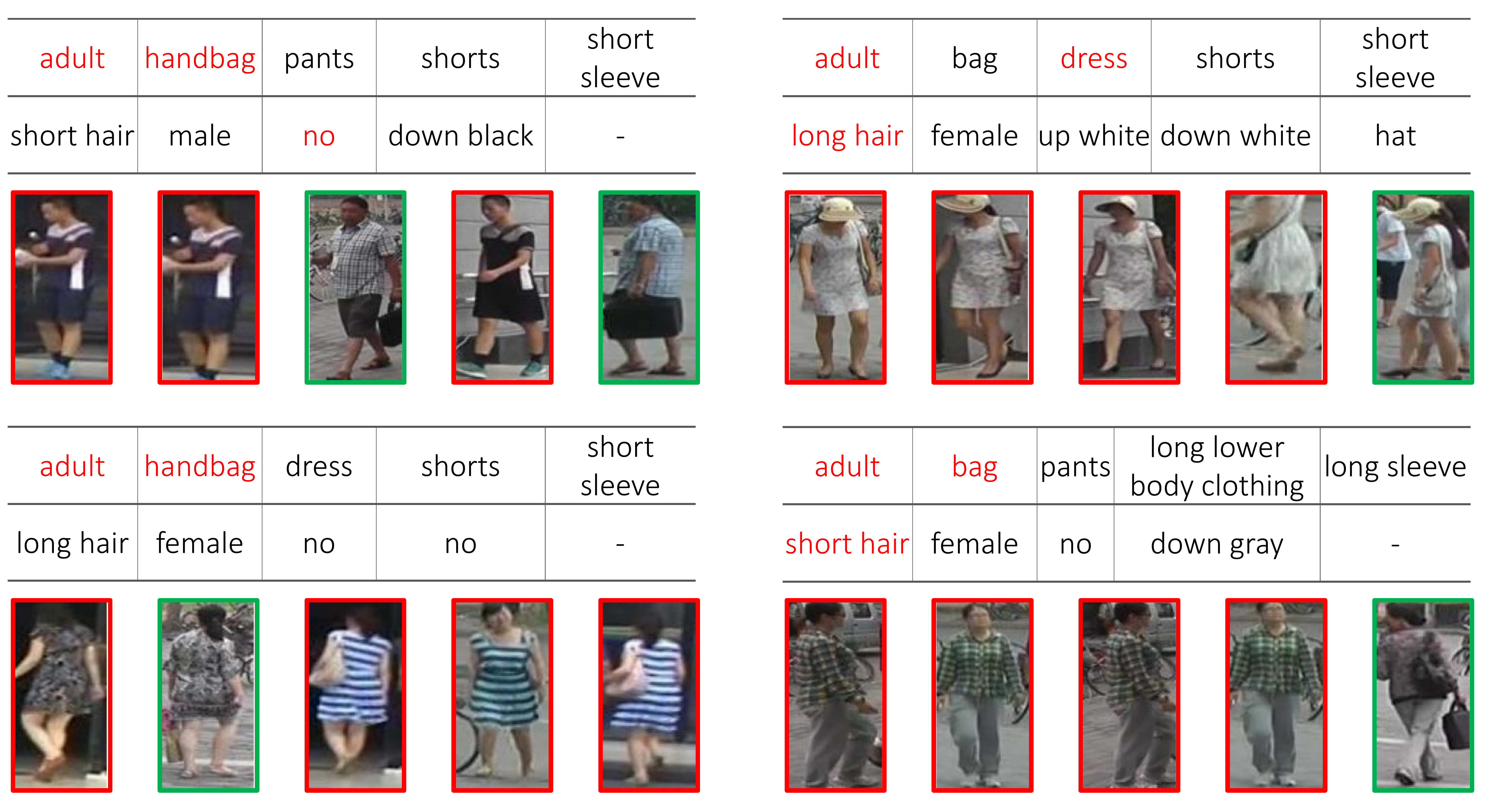}
    \caption{Failure cases of our method on the Market-1501 Attribute dataset. Images are sorted from left to right according to their ranks. Green and red boxes indicate true and false matches, respectively. Queries are presented above their retrieved images; blanks indicate attributes that do not exist in the query. Colored red in query indicates attributes that are different between query person category and person categort of false matches.
    }
    \label{fig:Failure}
    \vspace{-2mm}
\end{figure*}

Finally, in Fig.~\ref{fig:tsne_PETA},~\ref{fig:tsne_PA100K}, and~\ref{fig:tsne_Market}, we visualize the embedding manifold learned by our method through $t$-SNE on the test splits of the three datasets.
The visualization results demonstrate that for most images their nearest neighbors are similar with them in terms of their appearance traits. 
This suggests that our method learns a semantic relation between images and person categories successfully through the proposed loss.

\begin{figure*}[t!]
    \centering
    \includegraphics[width=\textwidth]{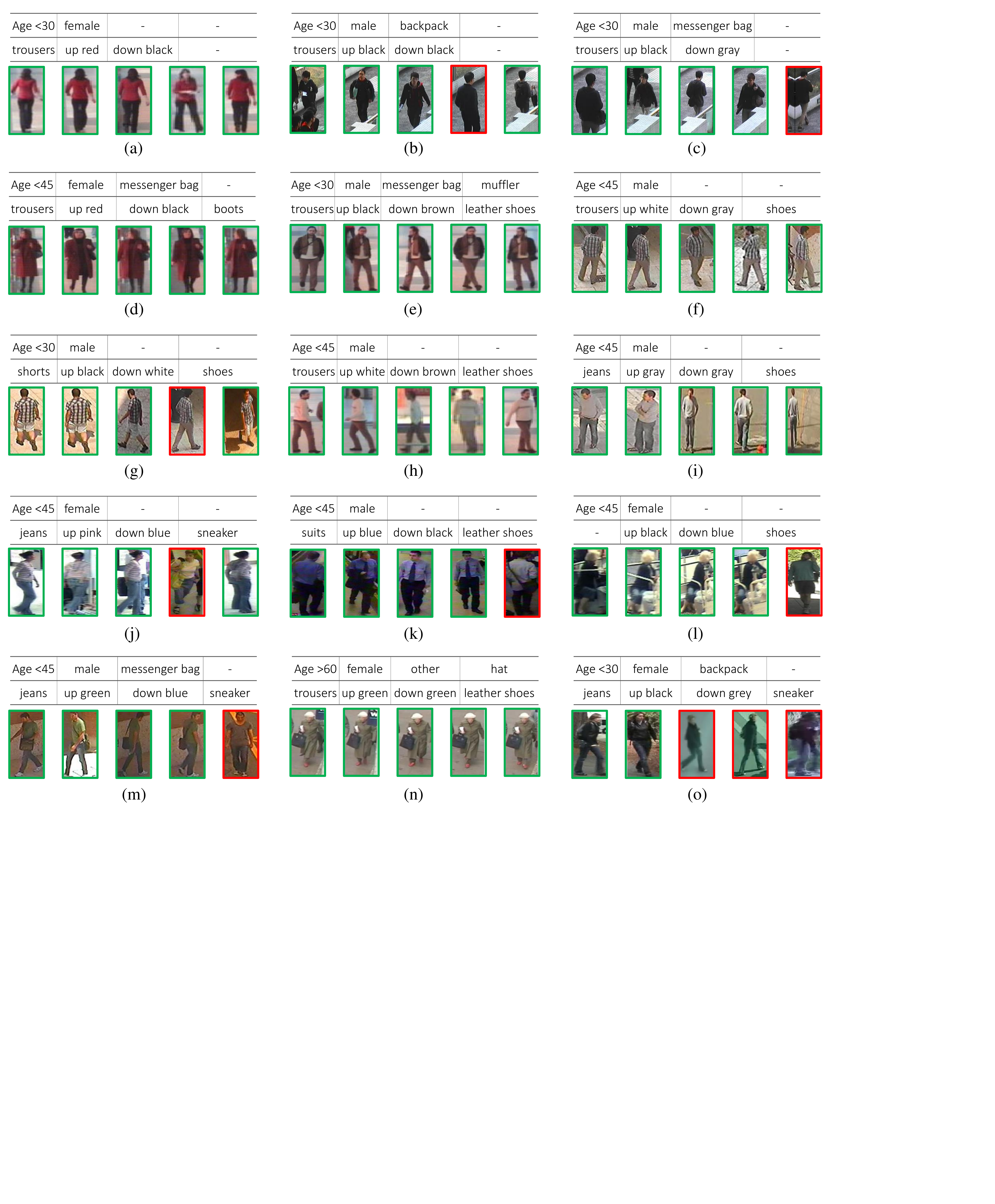}
    \vspace{-5mm}
    \caption{
    Top 5 retrieval results of our method on the PETA dataset. Images are sorted from left to right according to their ranks. Green and red boxes indicate true and false matches, respectively. Queries are presented above their retrieved images; blanks indicate attributes that do not exist in the query.
    }
    \label{fig:Supple_Qual_PETA}
    \vspace{1mm}
\end{figure*}

\begin{figure*}[h!]
    \centering
    \includegraphics[width=\textwidth]{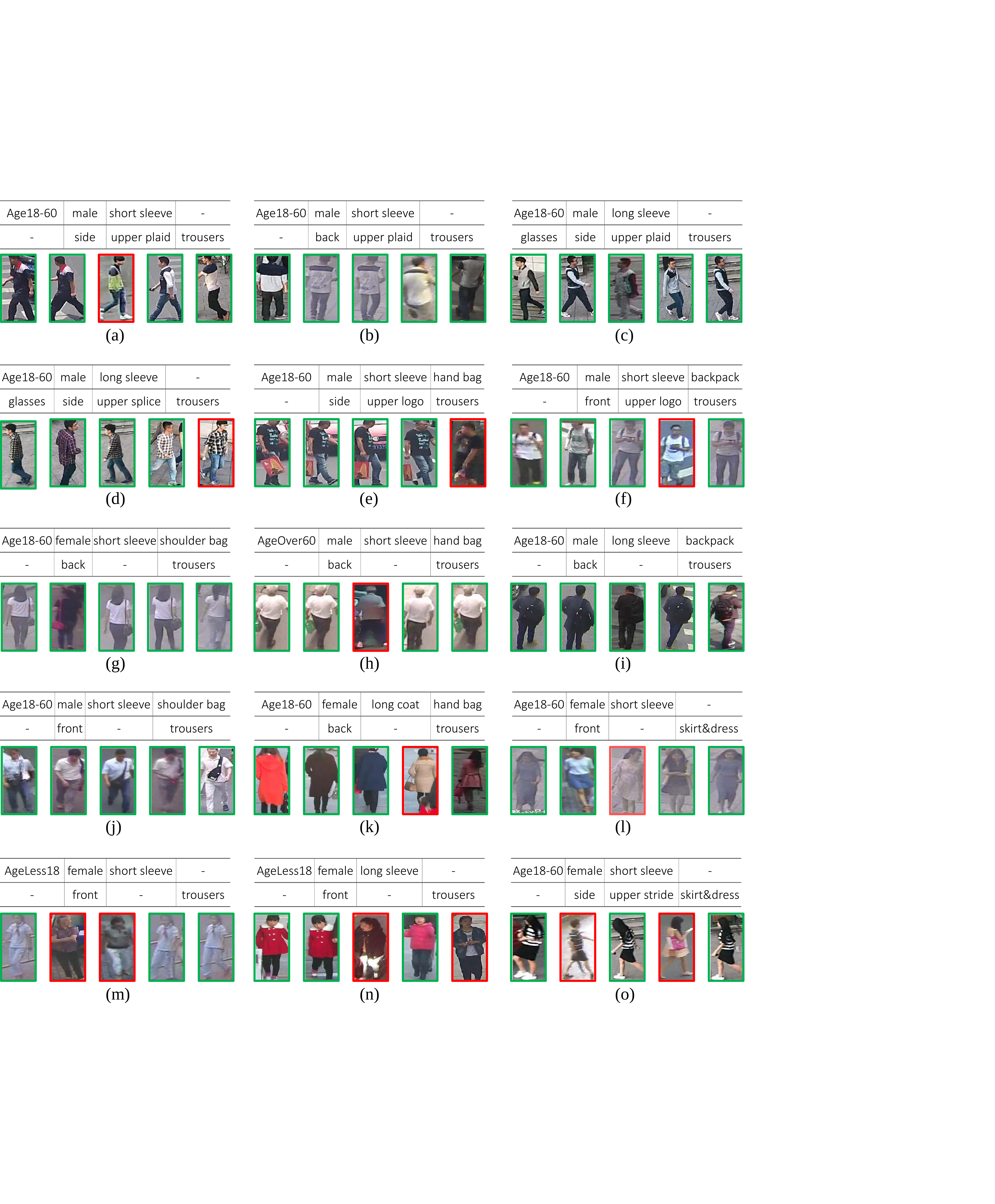}
    \vspace{-5mm}
    \caption{
    Top 5 retrieval results of our method on the PA100K dataset. Images are sorted from left to right according to their ranks. Green and red boxes indicate true and false matches, respectively. Queries are presented above their retrieved images; blanks indicate attributes that do not exist in the query.
    }
    \label{fig:Supple_Qual_PA100K}
    \vspace{-2mm}
\end{figure*}

\begin{figure*}[t!]
    \centering
    \includegraphics[width=\textwidth]{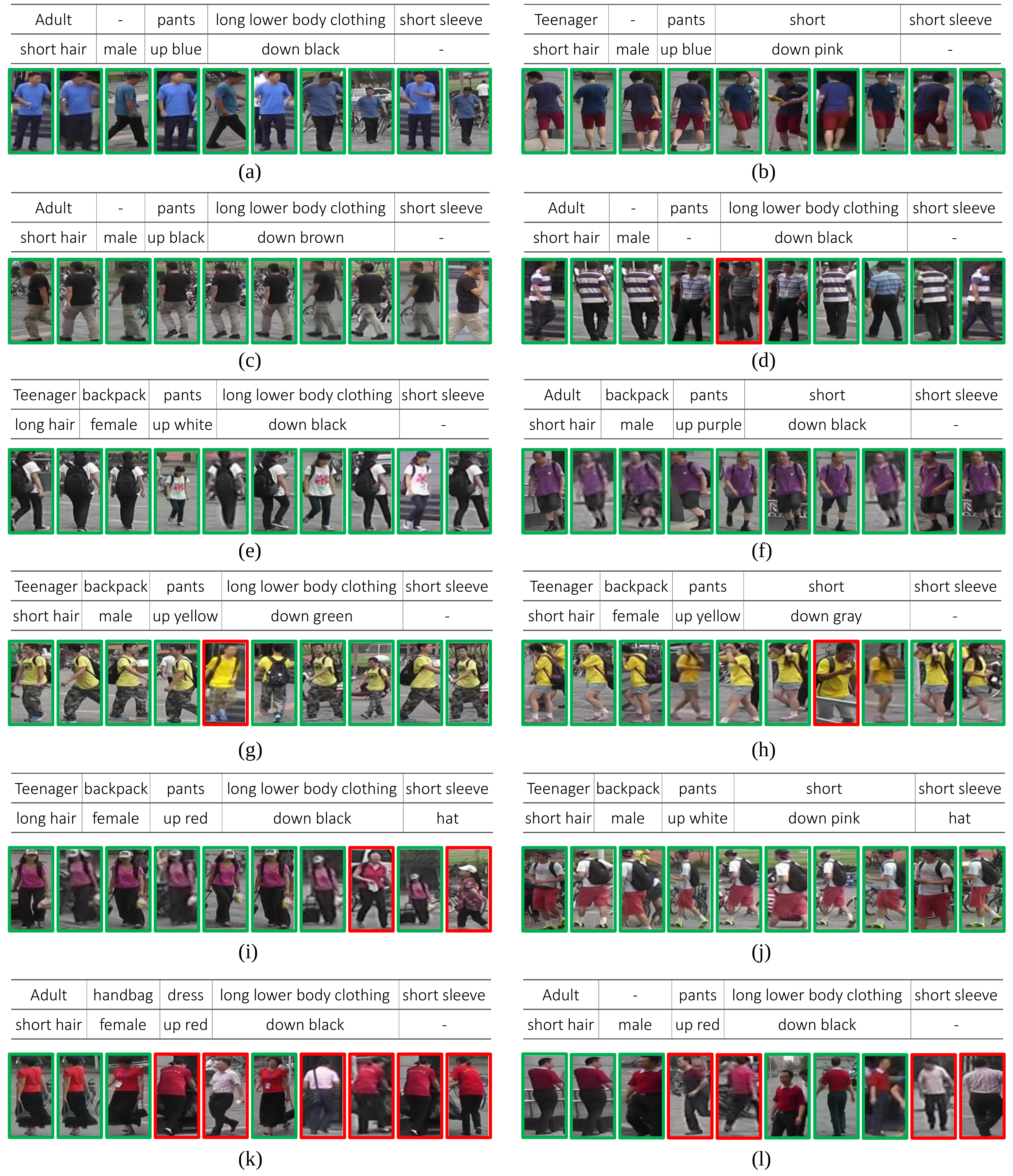}
    \caption{
    Top 10 retrieval results of our method on the Market-1501 Attribute dataset. Images are sorted from left to right according to their ranks. Green and red boxes indicate true and false matches, respectively. Queries are presented above their retrieved images; blanks indicate attributes that do not exist in the query.
    }
    \label{fig:Supple_Qual_Market}
\end{figure*}

\begin{figure*}[ht!]
    \centering
    \includegraphics[width=\textwidth]{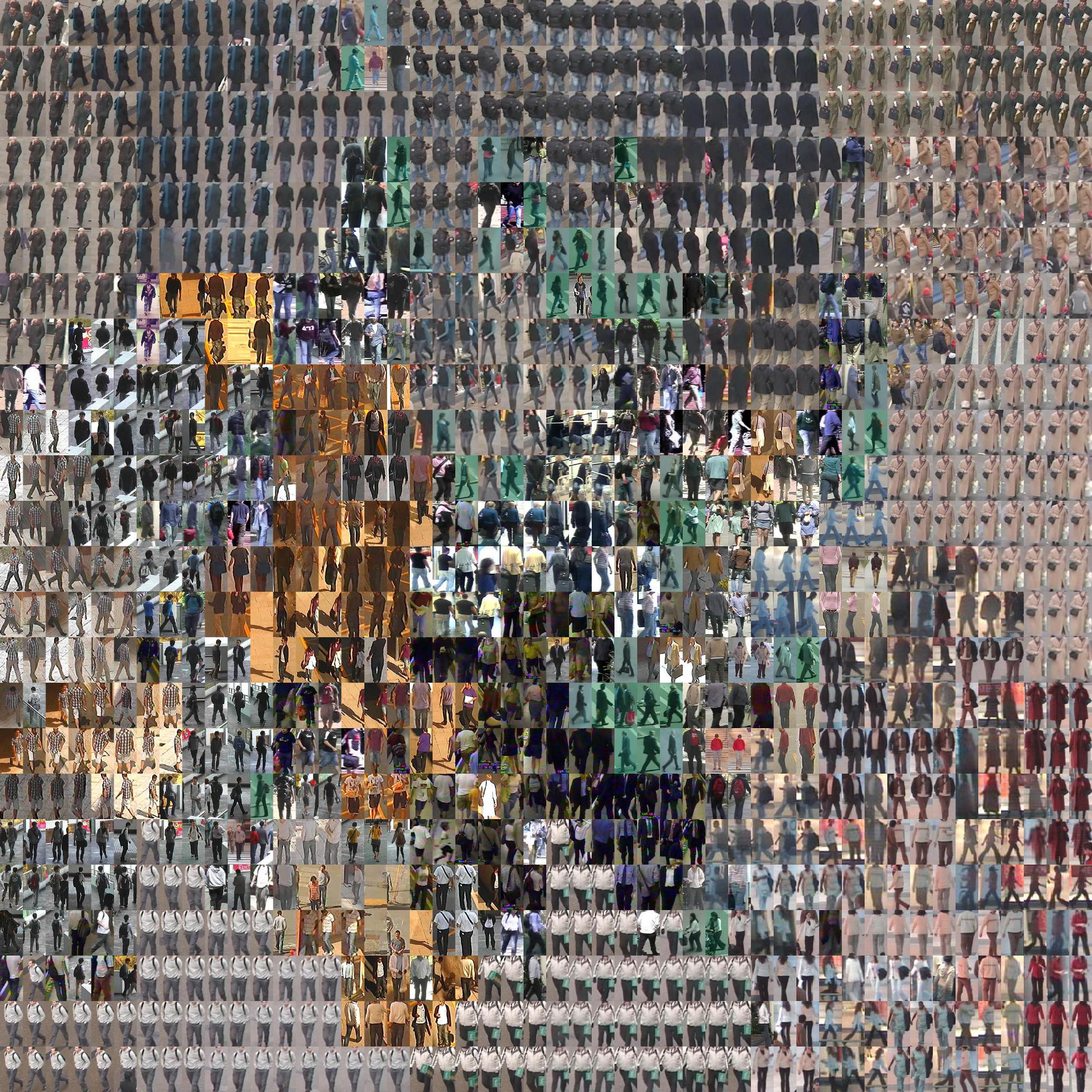}
    \caption{
    2D $t$-SNE visualization of image embeddings in the gallery of PETA.
    }
    \label{fig:tsne_PETA}
\end{figure*}
\begin{figure*}[ht!]
    \centering
    \includegraphics[width=\textwidth]{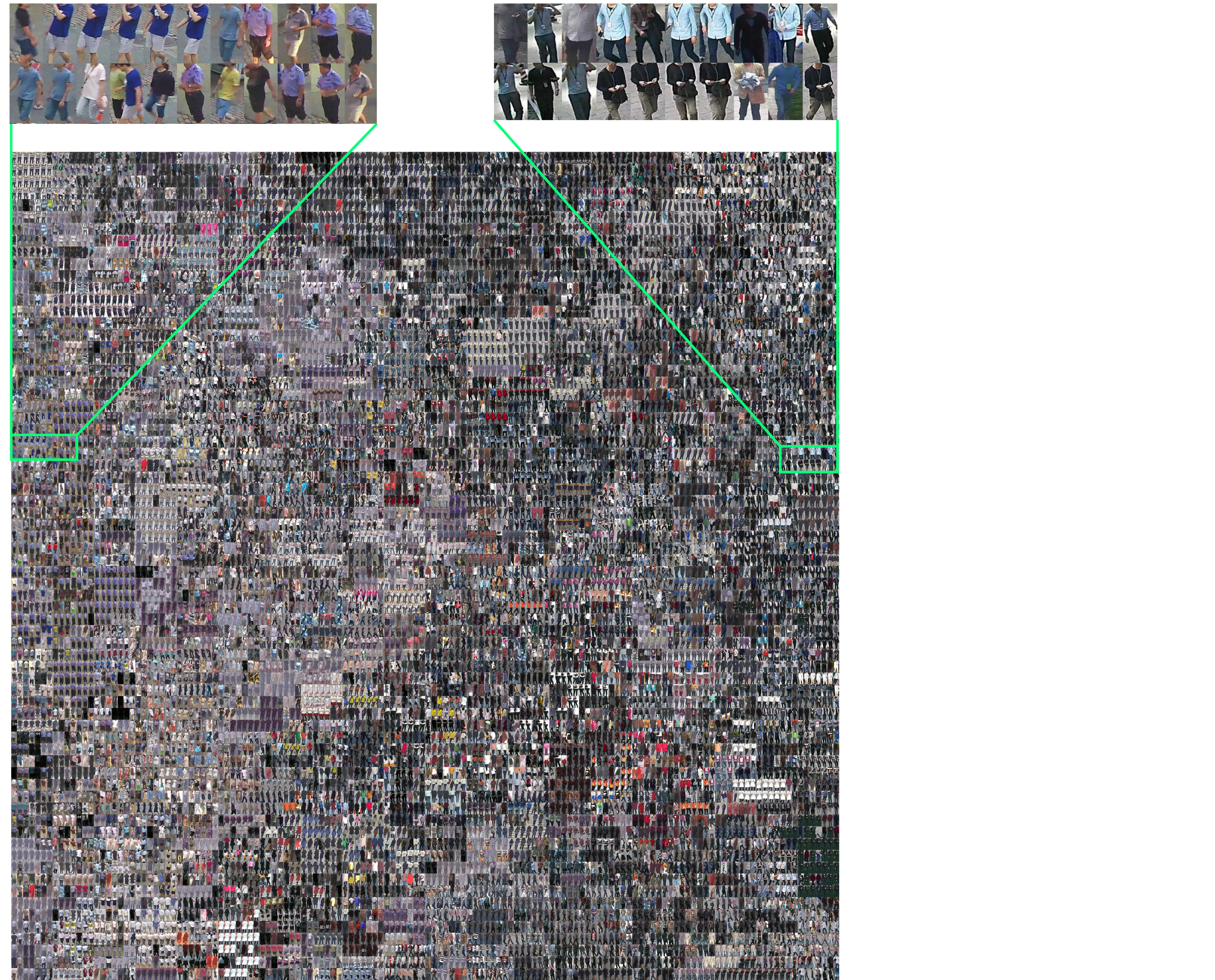}
    \caption{
    2D $t$-SNE visualization of image embeddings in the gallery of PA100K.
    }
    \label{fig:tsne_PA100K}
\end{figure*}
\begin{figure*}[ht!]
    \centering
    \includegraphics[width=\textwidth]{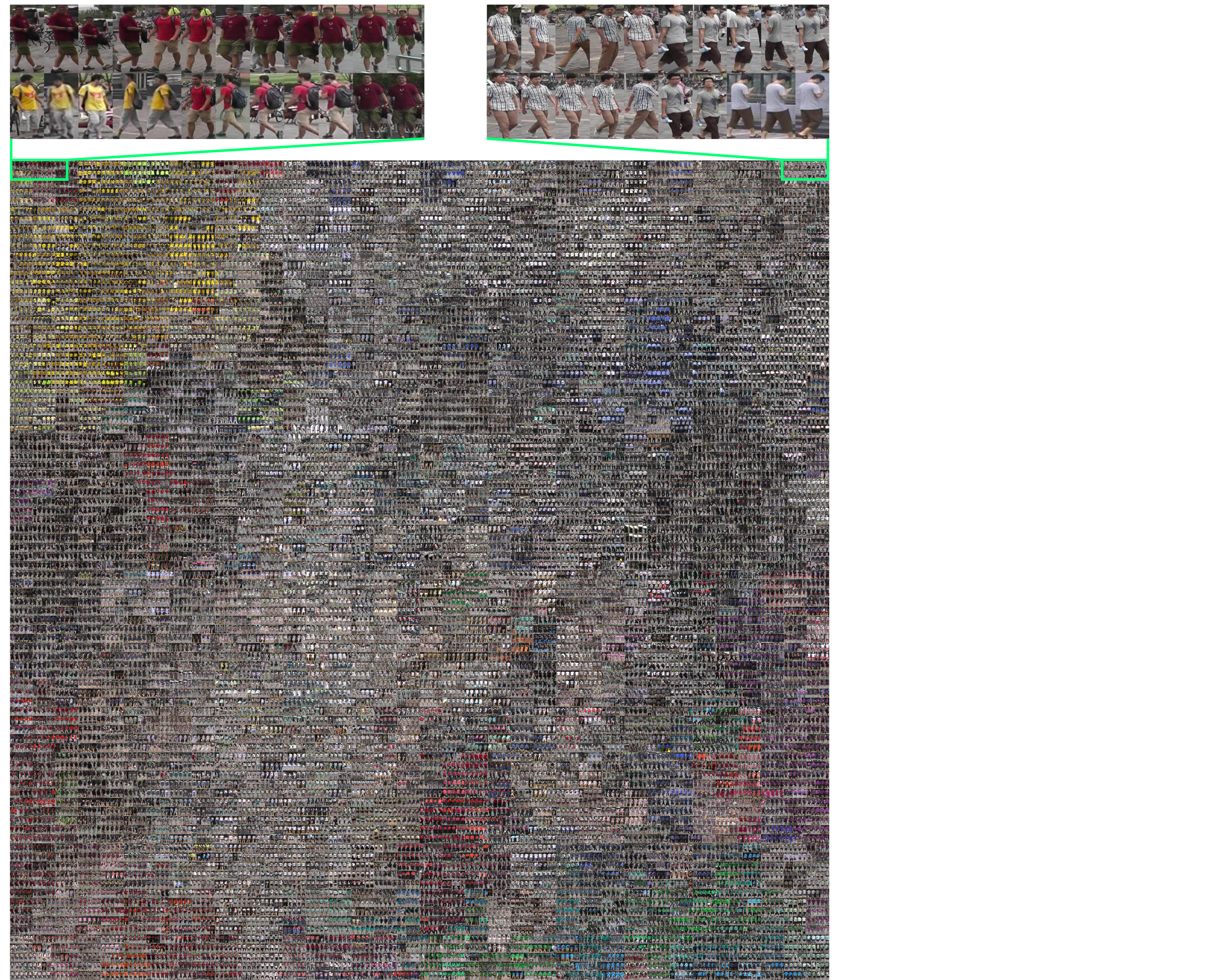}
    \caption{
    2D $t$-SNE visualization of image embeddings in the gallery of Market-1501 Attribute.
    }
    \label{fig:tsne_Market}
\end{figure*}
\newpage

\end{document}